\documentclass[journal]{IEEEtran}
\usepackage{graphicx}
\usepackage{multirow}
\usepackage{epstopdf}
\usepackage{times}
\usepackage{amsmath}
\usepackage{amssymb}
\usepackage{amsfonts}
\usepackage{color}
\usepackage{subfigure}
\usepackage{hyperref}

\usepackage{algorithm}
\usepackage{algorithmic}
\usepackage{rotating}

\usepackage{adjustbox,lipsum}
\usepackage{dsfont}
\usepackage{booktabs}
\usepackage{multirow}
\usepackage{array}
\usepackage{graphicx}

\newcommand{\eg}{{\em e.g.}}           
\newcommand{\ie}{{\em i.e.}}           

\begin{document}

\title{Adversarial Camera Alignment Network for Unsupervised Cross-camera Person Re-identification}

\author{Lei Qi,
        Lei Wang,
        Jing Huo,
        Yinghuan Shi,
        Xin Geng$^*$,
        Yang Gao

\thanks{This work was supported by the National Key Research and Development Plan of China (No. 2017YFB1002801), Project funded by China Postdoctoral Science Foundation (2021M690609), the National Science Foundation of China (62076063, 61806092) and the Jiangsu Natural Science Foundation (BK2018032). (Corresponding author: Xin Geng.)}
\thanks{Lei Qi and Xin Geng are with the School of Computer Science and Engineering, and the Key Lab of Computer Network and Information Integration (Ministry of Education), Southeast University, Nanjing, China, 211189 (e-mail: qilei@seu.edu.cn; xgeng@seu.edu.cn).}
\thanks{Lei Wang is School of Computing and Information Technology, University of Wollongong, Wollongong, Australia (e-mail: leiw@uow.edu.au).}
\thanks{Jing Huo, Yinghuan Shi and Yang Gao are with the State Key Laboratory for Novel Software Technology, Nanjing University, Nanjing, China, 210023 (e-mail: huojing@nju.edu.cn; syh@nju.edu.cn; gaoy@nju.edu.cn).}
\thanks{Part of this work was done when Lei Qi stayed at the State Key Laboratory for Novel Software Technology, Nanjing University.}

}

%
%

\markboth{~}%
{Shell \MakeLowercase{\textit{et al.}}: Bare Demo of IEEEtran.cls for IEEE Journals}

\maketitle

\begin{abstract}
In person re-identification (Re-ID), supervised methods usually need a large amount of expensive label information, while unsupervised ones are still unable to deliver satisfactory identification performance.
 In this paper, we introduce a novel person Re-ID task called unsupervised cross-camera person Re-ID, which only needs the within-camera (intra-camera) label information but not cross-camera (inter-camera) labels which are more expensive to obtain. In real-world applications, the intra-camera label information can be easily captured by tracking algorithms and few manual annotations. In this situation, the main challenge becomes the distribution discrepancy across different camera views, caused by the various body pose, occlusion, image resolution, illumination conditions, and background noises in different cameras. To address this situation, we propose a novel Adversarial Camera Alignment Network (ACAN) for unsupervised cross-camera person Re-ID. It consists of the camera-alignment task and the supervised within-camera learning task. To achieve the camera alignment, we develop a Multi-Camera Adversarial Learning (MCAL) to map images of different cameras into a shared subspace. Particularly, we investigate two different schemes, including the existing GRL (i.e., gradient reversal layer) scheme and the proposed scheme called ``other camera equiprobability'' (OCE), to conduct the multi-camera adversarial task.
Based on this shared subspace, we then leverage the within-camera labels to train the network. 
Extensive experiments on five large-scale datasets demonstrate the superiority of ACAN over multiple state-of-the-art unsupervised methods that take advantage of labeled source domains and generated images by GAN-based models. In particular, we verify that the proposed multi-camera adversarial task does contribute to the significant improvement.
\end{abstract}


%
\IEEEpeerreviewmaketitle

\section{Introduction}
\IEEEPARstart{P}{erson} re-identification (Re-ID) is to match images of the same individual captured by different cameras with non-overlapping camera views~\cite{DBLP:journals/tmm/ChenLLCH11}. Broadly, person Re-ID can be treated as a special case of the image retrieval problem with the goal of querying from a large-scale gallery set to quickly and accurately find the images that match a probe image~\cite{DBLP:journals/tip/ZhangLZZZ15}\cite{DBLP:journals/tmm/Wang0TL13}.
In recent years, person Re-ID has drawn an increasing interest in both academia and industry due to its great potentials in video surveillance applications~\cite{DBLP:journals/tifs/MaJZTP20,DBLP:journals/tifs/YeL0Y20}. 

\begin{figure}
\centering
\includegraphics[width=8.5cm]{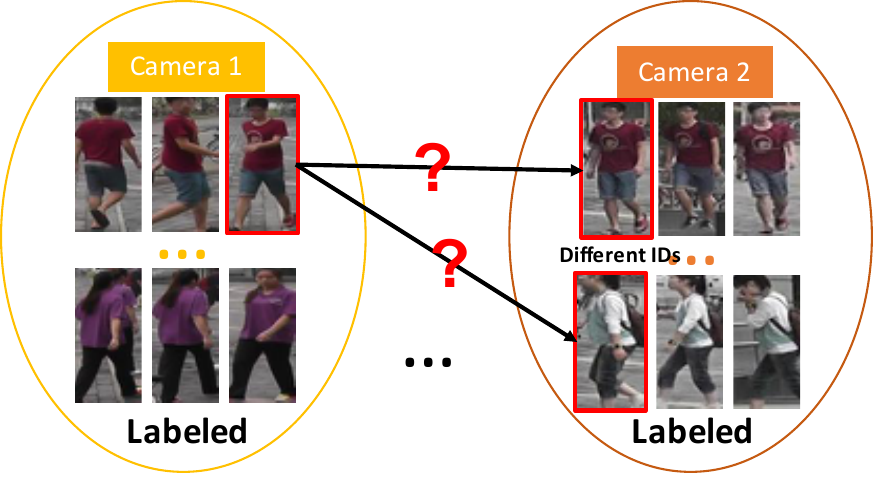}
\caption{The setting of unsupervised ``cross-camera'' person Re-ID. Note that we only show two cameras in this figure. As seen, we know the relationship of any two images from the same camera, while we do not have the label information of any two images from different cameras.}
\label{fig10}
\vspace*{-20pt}
\end{figure}

Currently, most existing work for person Re-ID mainly focuses on the supervised~\cite{DBLP:conf/cvpr/XiaoLOW16,DBLP:conf/aaai/ChenCZH17,sun2018beyond,zheng2018re,zheng2019pose,wu2019cross} and unsupervised~\cite{fan2018unsupervised,zhong2018generalizing,Bak_2018_ECCV,wang2018transferable,lv2018unsupervised} cases. Although the supervised person Re-ID methods can achieve good performance in many public datasets, they need a large-scale labeled dataset to train models, especially for the deep-learning-based methods. However, labeling data incurs a significant cost, especially when the identities across different camera views need to be matched. Therefore, supervised methods usually do not scale well in real-world applications. On the other hand, for the unsupervised Re-ID methods, due to the complete lack of label information to train sufficiently good models, they still have a big performance gap with their supervised counterparts. 

In this paper, instead of purely focusing on either supervised or unsupervised setting, we propose an \textbf{U}nsupervised ``\textbf{C}ross-\text{C}amera'' person Re-ID task (\textbf{UCC} Re-ID task), 
which only has the intra-camera (within-camera) labels but no inter-camera (cross-camera) label information, as illustrated in Fig.~\ref{fig10}. Note that in practice, it is much easier to obtain the within-camera label information by employing tracking algorithms~\cite{dehghan2015gmmcp,maksai2017non} and conducting a small amount of manual labeling. Therefore, this work aims to well exploit this new person Re-ID setting. In other words, we aim to utilize the intra-camera labeled data to effectively improve the performance of person Re-ID, thus substantially reducing the gap between the unsupervised Re-ID and the supervised ones.

For this proposed UCC person Re-ID setting, the main issue is the lack of cross-camera label information. This could result in poor generalization ability across camera views if merely using the within-camera labeled data to train models. In the literature, to deal with this problem, several unsupervised person Re-ID methods~\cite{fan2018unsupervised,zhong2018generalizing,Bak_2018_ECCV,wang2018transferable,lv2018unsupervised} generate pseudo-labels for unlabeled data. Moreover, in~\cite{DBLP:conf/bmvc/LinLLK18,DBLP:journals/corr/abs-1904-01308,qi2019novel}, other labeled datasets (e.g., source domains) are utilized to enhance the generalization ability of models in the target domain. Differently, in our UCC Re-ID case, we do not need any other labeled datasets to help to train the model. Instead, we focus on exploring the underlying information within the data via advanced techniques such as adversarial learning. 

\begin{figure}
\centering
\subfigure[Market1501]{
\includegraphics[width=3.5cm]{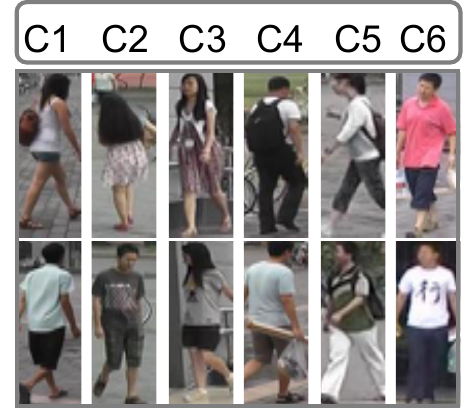}
}
\subfigure[DukeMTMC-reID]{
\includegraphics[width=4.67cm]{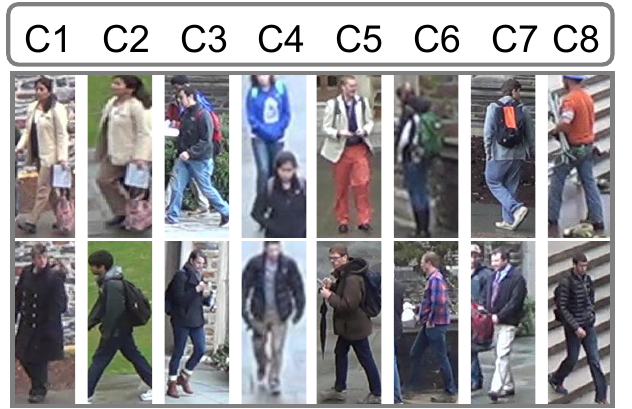}
}
\caption{Images from Market1501 and DukeMTMC-reID. (a) Six different cameras on Market1501; (b) Eight different cameras on DukeMTMC-reID. In the figure, C$i$ represents the $i$-th camera. As seen in this figure, there are varying view angle, occlusion, backgrounds, etc. in different cameras.}
\label{fig2}
\vspace*{-10pt}
\end{figure}

\begin{figure}
\centering
\subfigure[Market1501]{
\includegraphics[width=4cm]{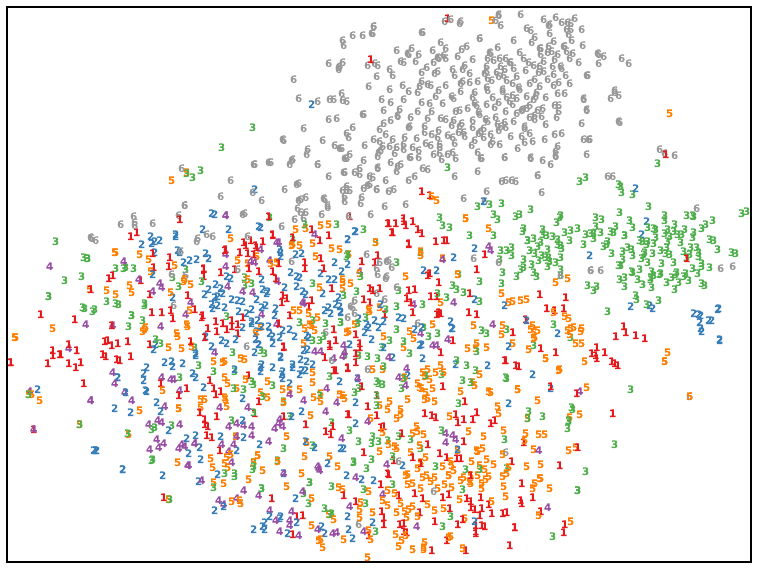}
}
\subfigure[DukeMTMC-reID]{
\includegraphics[width=4cm]{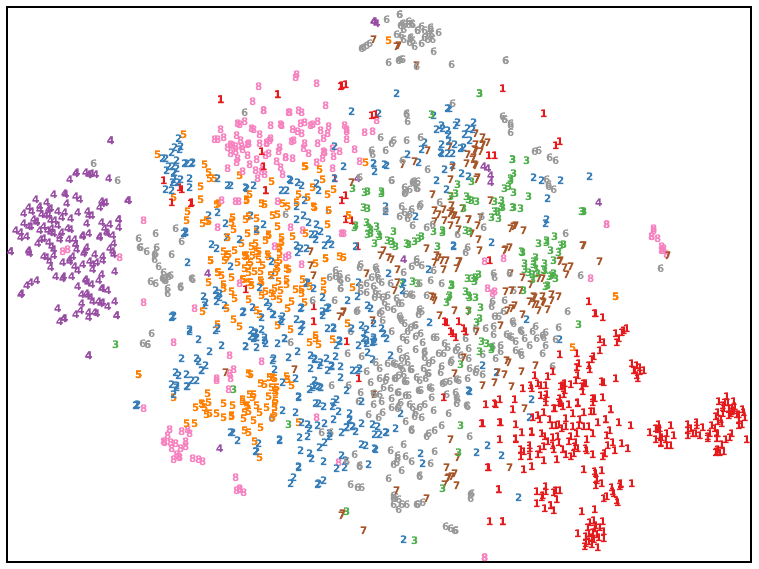}
}
\caption{Visualization of the distributions of two datasets via t-SNE~\cite{maaten2008visualizing}. The features of all images are extracted by the ResNet50~\cite{DBLP:conf/cvpr/HeZRS16} pre-trained on ImageNet~\cite{DBLP:conf/cvpr/DengDSLL009}. Different colors denote the images from different cameras. In detail, (a) shows six different cameras on Market1501; (b) shows eight different cameras on DukeMTMC-reID. Best viewed in color.}
\label{fig1}
\vspace*{-15pt}
\end{figure}

In real-world scenarios, a person's appearance often varies greatly across camera views due to changes in body pose, view angle, occlusion, image resolution, illumination conditions, and background noises, as shown in Fig.~\ref{fig2}. These variations lead to the data distribution discrepancy across cameras. To better illustrate this situation, we use the ResNet50~\cite{DBLP:conf/cvpr/HeZRS16} pre-trained on ImageNet~\cite{DBLP:conf/cvpr/DengDSLL009} to extract features from raw images and employ the t-SNE~\cite{maaten2008visualizing} to visualize the data distribution on the benchmark datasets of Market1501 and DukeMTMC-reID in Fig.~\ref{fig1}. As seen, samples of different cameras usually reside in different regions, i.e., the involved cameras do not align with each other in this feature space. This is the key challenge to deal with in the proposed UCC person Re-ID task. Particularly, if all samples from different camera views could conform to the same data distribution in a shared feature space, we will obtain sufficiently good performance by simply using the within-camera label information to train models.

To deal with the above distribution discrepancy across camera views, we propose a novel Adversarial Camera Alignment Network (ACAN) for the proposed
UCC person re-identification task. Specifically, we develop a Multi-Camera Adversarial Learning (MCAL) method to project all data from different cameras into a common feature space. Then we utilize the intra-camera label information to conduct the discrimination task that aims to make the same identities closer and different identities farther. To carry out the multi-camera adversarial learning, we develop two different schemes to maximize the loss function of the discriminator (i.e., a classifier to discriminate data from different cameras) in the ACAN framework including the widely used gradient reversal layer (GRL) and a new one called ``other camera equiprobability'' (OCE) proposed by this work. In particular, we also give the theoretical analysis to show that cameras can be completely aligned when the discriminator predicts an equal probability for each camera class. Extensive experiments on five large-scale datasets including three image datasets and two video datasets well validate the superiority of the proposed work when compared with the state-of-the-art unsupervised Re-ID methods utilizing the labeled source domains and generated images by the GAN-based models. Furthermore, the experimental study shows that aligning the distributions of different cameras can indeed bring a significant improvement in the proposed framework.

In summary, the main contributions of this paper are fourfold. Firstly, we propose a new person Re-ID setting named Unsupervised Cross-Camera (UCC) person Re-ID which only needs the within-camera labels but not the cross-camera label information that is more expensive to collect. 
Secondly, we develop an adversarial camera alignment network for the proposed task, which deals with the cross-camera unlabeled case from the perspective of reducing the cross-camera data distribution discrepancy. 
Thirdly, to achieve the multi-camera adversarial learning, we utilize the existing gradient reversal layer (GRL) and also propose a new scheme called ``other camera equiprobability'' (OCE), with theoretical analysis provided for the latter. Lastly, the experimental results on multiple benchmark datasets show that the proposed method outperforms the state-of-the-art unsupervised methods in the UCC setting. Also, the efficacy of the proposed multi-camera adversarial learning method is well confirmed.

The rest of this paper is organized as follows.
The related work is reviewed in Section \ref{s-related}.
The ACAN framework is proposed and discussed in Section \ref{s-framework}.
Experimental results and analysis are presented in Section \ref{s-experiment},
and the conclusion is drawn in Section \ref{s-conclusion}.

\section{Related work}\label{s-related}

\subsection{Unsupervised Person Re-ID}
In the early years, most methods for person Re-ID mainly focus on designing the discriminative hand-crafted features containing the color and structure information, such as LOMO~\cite{liao2015person} and BoW~\cite{DBLP:conf/iccv/ZhengSTWWT15}, which can be directly applied to person Re-ID without using any label information. Besides, some domain adaptation methods~\cite{DBLP:conf/cvpr/PengXWPGHT16, DBLP:journals/tcsv/WangZLZ16, DBLP:journals/tip/MaLYL15,qi2018unsupervised,DBLP:conf/iccv/YuWZ17} based on hand-crafted features are developed to transfer the discriminative information from a labeled source domain to an unlabeled target domain by learning a shared subspace or dictionary between source and target domains. However, these methods are not based on deep learning and thus do not fully explore the high-level semantics in images.

With the introduction of deep learning into the computer vision community, many methods based on deep learning have been developed for Re-ID. Since deep networks need the label information to train, many methods generate pseudo labels for unlabeled samples~\cite{fan2018unsupervised,zhong2018generalizing,Bak_2018_ECCV,wang2018transferable,lv2018unsupervised}. In~\cite{fan2018unsupervised}, clustering methods are utilized to assign pseudo-labels to unlabeled samples. Lv \textit{et al.}~\cite{lv2018unsupervised} leverage spatio-temporal patterns of pedestrians to obtain robust pseudo-labels. In~\cite{wang2018transferable}, an approach is proposed to simultaneously learn an attribute-semantic and identity-discriminative feature representation by producing the pseudo-attribute-labels in target domain. However, generating pseudo-labels for unlabeled data is complex, and they may not be necessarily consistent with true labels.

In addition, deep domain adaptation methods are developed to reduce the discrepancy between source and target domains from the perspective of feature representation~\cite{DBLP:conf/bmvc/LinLLK18,DBLP:journals/corr/abs-1904-01308,qi2019novel}.
Lin \textit{et al.}\cite{DBLP:conf/bmvc/LinLLK18} develop a novel unsupervised Multi-task Mid-level Feature Alignment (MMFA) network, which uses the Maximum Mean Discrepancy (MMD) to reduce the domain discrepancy.
Delorme \emph{et al.}~\cite{DBLP:journals/corr/abs-1904-01308} introduce an adversarial framework in which the discrepancy across cameras is relieved by fooling a camera discriminator.
Considering the presence of camera-level sub-domains in person Re-ID,
Qi \emph{et al.}~\cite{qi2019novel} develop a camera-aware domain adaptation to reduce the discrepancy not only between source and target domains but also across these sub-domains (i.e., cameras).

Recently, 
generating extra training images for target domain has become popular~\cite{wei2018person,deng2018image,zhong2018generalizing,Bak_2018_ECCV}. 
Wei \textit{et al.}~\cite{wei2018person} impose constraints to maintain the identity in image generation. The approach in~\cite{deng2018image} enforces the self-similarity of an image before and after translation and the domain-dissimilarity of a translated source image and a target image. Zhong \textit{et al.}~\cite{zhong2018generalizing} propose to seek camera-invariance by using unlabeled target images and their camera-style transferred counterparts as positive match pairs. Besides, it views the source and target images as negative pairs for the domain connectedness. 
In~\cite{zhong2019invariance}, the camera-invariance is introduced into the model, which requires that each real image and its style-transferred counterparts share the same identity.

Besides the aforementioned image-based unsupervised methods, several video-based unsupervised methods for Re-ID have also been seen in recent years~\cite{kodirov2016person,khan2016unsupervised,ye2018robust,liu2017stepwise,ye2017dynamic}. 
To generate robust cross-camera labels, Ye~\emph{et al.}~\cite{ye2017dynamic} construct a graph for the samples in each camera, and then introduce a graph matching scheme for the cross-camera label association. 
Chen \emph{et al.}~\cite{DBLP:conf/bmvc/ChenZG18} learn a deep Re-ID matching model by jointly optimizing two margin-based association losses in an end-to-end manner, which effectively constrains the association of each frame to the best-matched intra-camera representation and cross-camera representation.
Li \emph{et al.}~\cite{Li_2018_ECCV} jointly learn within-camera tracklet association and cross-camera tracklet correlation by maximizing the discovery of most likely tracklet relationships across camera views. Moreover, its extension is capable of incrementally discovering and exploiting the underlying Re-ID discriminative information from automatically generated person tracklet data end-to-end~\cite{li2019unsupervised}. 

Different from all of the above methods, the proposed method in this work does not use any additional data, such as generated images from GANs or labeled source domains. Also, we do not need any complex pseudo-label schemes to generate the association across cameras.

\subsection{Unsupervised Domain Adaptation}
Unsupervised domain adaptation is also related to our work. It is a more general technique to reduce the distribution discrepancy between source and target domains.
In the literature, most unsupervised domain adaptation methods learn a common mapping between source and target distributions. 
Several methods based on the Maximum Mean Discrepancy (MMD) have been proposed~\cite{DBLP:conf/icml/LongC0J15,NIPS2016_6110,DBLP:journals/corr/ZhangYCW15,DBLP:journals/corr/TzengHZSD14}.
Long \textit{et al.}~\cite{DBLP:conf/icml/LongC0J15}  introduce a new deep adaptation network, where the hidden representations of all task-specific layers are embedded in a Reproducing Kernel Hilbert space. To transfer a classifier from the source domain to the target domain, the work in~\cite{NIPS2016_6110} jointly learns adaptive classifiers between the two domains by a residual function.  In~\cite{DBLP:conf/eccv/GhifaryKZBL16,DBLP:conf/nips/BousmalisTSKE16}, autoencode-based methods are investigated to explore the discriminative information in target domain. Recently, adversarial learning~\cite{DBLP:conf/icml/GaninL15,DBLP:journals/corr/abs-1803-09210,DBLP:conf/cvpr/TzengHSD17} has been applied to domain adaptation. Ganin \textit{et al.}~\cite{DBLP:conf/icml/GaninL15} propose the gradient reversal layer (GRL) to pursue the same distribution between source and target domains. Inspired by generative adversarial networks (GANs),
Tzeng \textit{et al.}~\cite{DBLP:conf/cvpr/TzengHSD17}  leverage a GAN loss to match the data distributions of source and target domains. 
Nevertheless, different from the typical unsupervised domain adaptation case which only has two domains (i.e., one source domain and one target domain), our UCC person Re-ID task usually faces the distribution discrepancy across multiple domains (i.e., cameras).

In recent years, some multi-domain adversarial methods have been developed to solve multi-source domain adaptation~\cite{zhao2018adversarial,xu2018deep,schoenauer-sebag2018multidomain}. In~\cite{zhao2018adversarial,xu2018deep}, adversarial learning is employed in each pair of the source domain and the target domain to map all domains into a common feature space. Sebag \textit{et al.}~\cite{schoenauer-sebag2018multidomain} utilize the gradient reversal layer (GRL) to map multiple domains into the same space.  
However, the above methods deal with the adversarial task in the setting of ``one to many'' (i.e., one target domain and multiple source domains). Differently, the UCC Re-ID in this work needs to handle the adversarial task in the setting of ``many to many'' (i.e., multiple cameras to multiple cameras), which is more complicated and needs a new method. 
\section{Adversarial Camera Alignment Network}\label{s-framework}
\begin{figure*}
\centering
\includegraphics[width=16cm]{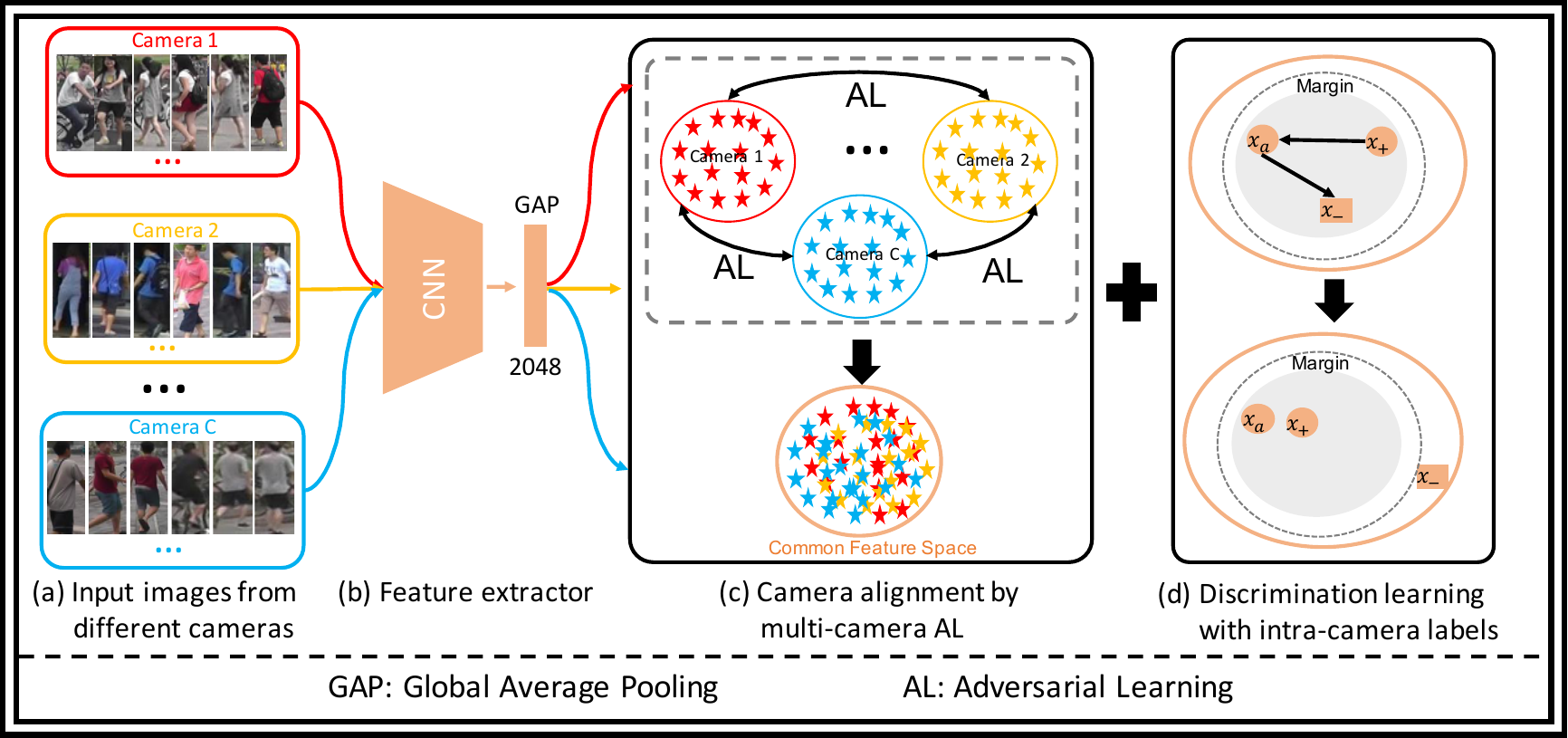}
\caption{An illustration of the proposed adversarial camera alignment network. It is an end-to-end framework. In detail, a) is the input images from different cameras, which only has intra-camera labels but not inter-camera label information; b) is the feature extractor to obtain the feature representation of each input image from the deep network; c) shows the camera-alignment task by adversarial learning to map all images of different cameras into a shared feature space; d) denotes the supervised intra-camera discrimination task with only using the intra-camera label information in this shared feature space, which aims to pull the images of the same identity closer and push the images of different identities farther. Best viewed in color.}
\label{fig3}
\vspace*{-20pt}
\end{figure*}
This section puts forward a novel Adversarial Camera Alignment Network (ACAN) for the unsupervised cross-camera (UCC) person Re-ID task focused in this paper. It only needs the within-camera (intra-camera) label information but no \textbf{cross-camera} (inter-camera) labels\footnote{The proposed unsupervised \textbf{cross-camera} person Re-ID task is also called \textit{UCC person Re-ID} in this paper.}. We illustrate the proposed method in Fig.~\ref{fig3}, which mainly consists of the camera alignment task and the discrimination learning task. In this paper, we develop a multi-camera adversarial learning method to align all cameras, with two different adversarial schemes to achieve this goal. For learning the discrimination information from the intra-camera labeled data, we simply employ the commonly used triplet loss with the hard-sample-mining scheme~\cite{hermans2017defense}. In the following part, we describe the camera alignment module and the discrimination learning module in Section~\ref{SEC:AC} and Section~\ref{SEC:LDI}, respectively. After that, the optimization of the objective function in the proposed ACAN is presented in Section~\ref{SEC:OPT}.
\subsection{Camera Alignment by Multi-camera Adversarial Learning} \label{SEC:AC}
In unsupervised cross-camera person Re-ID, we only have the within-camera label information and thus cannot directly explore the relationship between cross-camera images. Due to the variation of body pose, occlusion, image resolution, illumination conditions, and background noises, there exists significant data discrepancy across cameras, as shown in Fig.~\ref{fig1}. If we merely use the within-camera labels to train models, the data distribution discrepancy across cameras cannot be removed. Different from most existing unsupervised person Re-ID methods~\cite{fan2018unsupervised,zhong2018generalizing,Bak_2018_ECCV,wang2018transferable,lv2018unsupervised} which generate pseudo-labels for unlabeled data, we address this problem from the perspective of reducing data distribution. In other words, we reduce the distribution discrepancy by aligning all cameras with adversarial learning. To achieve the goal, we develop a Multi-Camera Adversarial Learning (MCAL) to map images of different cameras into a common feature space. 

Let $X=[X_{1}, ..., X_{C}]$ be the set of training images, where $X_{c}$ denotes the set of images from the $c$-th camera and $C$ is the total number of cameras. $Y=[Y_{1}, ..., Y_{C}]$ is the corresponding set of within-camera person labels. The set of camera IDs (i.e., the label of each camera class) of the images in $X$ is denoted by $Z$, i.e., each element in $Z$ represents the camera ID of each image. Adversarial learning involves the optimization of discriminator and generator~\cite{goodfellow2014generative}. As usual, the discriminator (i.e., a classifier to distinguish the images from different cameras) in this work is optimized by a cross-entropy loss defined on the $C$ camera classes as
\begin{equation}\label{eq11}
\begin{aligned}
&\min_{D}\mathcal{L}_\mathrm{MCAL-D}(X, Z, F)= \\
&\min_{D}\left[\mathbb{E}_{(x,z)\sim (X,Z)}\left(-\sum_{k=1}^{C}\delta(z-k)\log D(F(x), k)\right)\right],
  \end{aligned}
\end{equation} where $x$ denotes an image, $z$ is the true camera class label of $x$, and $\delta(\cdot)$ represents the Dirac delta function, \ie, it is equal to 1 if $k=z$, otherwise 0. $F$ denotes the backbone network (i.e., the feature extractor module in Fig.~\ref{fig3}), and $F(x)$ is the feature representation of $x$. $D$ indicates the discriminator and $D(F(x), k)$ is the prediction score (i.e., probability) for $x$ with respect to the $k$-th camera class.

To train the feature extractor $F$, we conduct the adversarial task so that the discriminator cannot effectively predict the camera ID of each image. Intuitively, to carry out the multi-camera adversarial task, we can directly optimize the feature extractor $F$ by maximizing the discriminator loss as
\begin{equation}\label{eq12}
\begin{aligned}
&\min_{F}\mathcal{L}_\mathrm{MCAL-F}(X, Z, D)\triangleq\max_{F}\mathcal{L}_\mathrm{MCAL-D}(X, Z, D) \\
&=\min_{F}\left[\mathbb{E}_{(x, z)\sim (X, Z)}\sum_{k=1}^{C}\delta(z-k)\log D(F(x), k)\right],
  \end{aligned}
\end{equation}where for consistency it is written as minimizing the negative discriminator loss. 

Following the literature, we first investigate the gradient reversal layer (GRL) technique~\cite{DBLP:conf/icml/GaninL15} to solve Eq.~(\ref{eq12}). After that, we point out its limitations for the multiple-camera adversarial task and develop a new simple but effective criterion ``other camera equiprobability'' (OCE) for our task. 

\subsubsection{GRL-based adversarial scheme}\label{SEC:AC-1}
The gradient reversal layer (GRL)~\cite{DBLP:conf/icml/GaninL15} is commonly used to reduce the distribution discrepancy of two domains. It is equal to maximizing the domain discrimination loss. From the perspective of our work, GRL can be viewed as maximizing the camera discrimination loss (i.e., Eq.~(\ref{eq12})). Therefore, we can utilize it to solve the multi-camera adversarial task. Note that because our case deals with multiple (camera) classes, a loss will be counted as long as an image is not classified into its true camera class. 

To train the feature extractor $F$ with Eq.~(\ref{eq12}), we insert GRL between $F$ and $D$ as in the literature~\cite{DBLP:conf/icml/GaninL15}. During forward propagation, GRL is simply an identity transform. During backpropagation, GRL reverses (i.e., multiplying by a negative constant) the gradients of the camera discriminator loss with respect to the network parameters in feature extraction layers and pass them backward. This GRL-based adversarial scheme can somehow work to reduce distribution discrepancy across different cameras (i.e., domains), as will be experimentally demonstrated shortly.  

However, we observe that this scheme has a drawback. Maximizing the discriminator loss by GRL only enforces an image ``not to be classified into its true camera class''. It will appear to be ``equivalently good'' for this optimization as long as an image is classified into any wrong camera classes. As a result, this scheme may be trapped into a ``local'' camera assignment (i.e., the images from a certain camera are only misclassified into a few cameras but not any other cameras). In the extreme case, this could lead to a ``many to one'' assignment (i.e., all images are misclassified into a single camera class, as shown in Fig.~\ref{fig8}), especially when the dataset is collected from many cameras. This adversely affects the reduction of data distribution discrepancy. The disadvantage of GRL will be further verified in Section~\ref{sec:EXP-FA}.
    
\subsubsection{OCE-based adversarial scheme}\label{SEC:AC-2}
To overcome the above issue, we explicitly assign camera class labels to each training image. To find a good scheme of the camera label assignment to train the feature extractor $F$, we give the following theoretical analysis. 

\textbf{Proposition 1.}
\textit{Given any image $x$, we explicitly assign it, with equal probability, all the C camera class labels. Training the feature extractor $F$ with this condition makes the data distribution of all cameras aligned as
\begin{equation*}
\begin{aligned}
KL(p(x|\mathcal{C}_1)\parallel p(x))=\cdots =KL(p(x|\mathcal{C}_C)\parallel p(x))=0,
\end{aligned}
\end{equation*} where $p(x|\mathcal{C}_i)$ is the class-conditional probability density function of the $i$-th camera class, $p(x)$ denotes the probability density function of the images. $KL(p(x|\mathcal{C}_i)\parallel p(x))$ is the Kullback--Leibler divergence between $p(x|\mathcal{C}_i)$ and $p(x)$.}

\textit{Proof.}
 Given an image $x$, its posteriori probability with respect to the $i$-th camera class (denoted by $\mathcal{C}_i (i=1,\cdots,C)$) can be expressed via the Bayes' rule as 
\begin{equation}\label{eqn:Bayes}
P(\mathcal{C}_i|x) = \frac{p(x|\mathcal{C}_i)P(\mathcal{C}_i)}{p(x)}, \quad i=1,\cdots,C, 
\end{equation}where $p(x|\mathcal{C}_i)$ indicates the class-conditional probability density function of the $i$-th camera class, $p(x)$ is the probability density function of the images, and $P(\mathcal{C}_i)$ is the priori probability of the $i$-th camera class. Note that $P(\mathcal{C}_i|x)$ is just $D(F(x),i)$. For clarity, $D(F(x),i)$ is compactly denoted by $D_i$.

For an image $x$, if we assign the equal label (i.e., $1/C$) to all camera classes to train the feature extractor $F$, we can get the loss function as
\begin{equation}\label{eq22}
\underset{\{D_1,\cdots,D_{C}\}}\min\left( -\frac{1}{C}\sum_{i=1}^{C}\log D_i  \right)
\end{equation} with the constraints of $D_i\geq{0}$ and $\sum_{i=1}^{C}D_i=1$. Due to the symmetry of the objective function with respect to the probabilities $D_1,\cdots,D_{C}$, it is not difficult to see that the optimal value of $D_i$ is $1/C$ for $i=1,\cdots,C$. A rigorous proof can be readily obtained by applying the Karush-Kuhn-Tucker conditions~\cite{boyd2004convex} to this optimization. This indicates that $P(\mathcal{C}_i|x)$ will equal $1/C$ when Eq.~(\ref{eq22}) is minimized for this given image $x$.  

We turn to Eq. (\ref{eqn:Bayes}) and rewrite it as
\begin{equation}\label{eqn:Bayes-1}
p(x|\mathcal{C}_i) = \frac{P(\mathcal{C}_i|x)}{P(\mathcal{C}_i)}p(x),\quad i=1,\cdots,C. 
\end{equation}Without loss of generality, equal priori probability can be set for the $C$ camera classes, that is, $P(\mathcal{C}_i)$ is constant $1/C$. Further, note that by optimizing $D_i$ in Eq.~(\ref{eq22}) above, it can be known that 
\begin{equation}\label{eqn:Bayes-1-1}
P(\mathcal{C}_i|x)=\frac{1}{C},\quad i=1,\cdots,C.  
\end{equation}
Combining the above results, Eq.~(\ref{eqn:Bayes-1}) becomes
\begin{equation}\label{eqn:Bayes-2}
p(x|\mathcal{C}_i) = \frac{1/C}{1/C}p(x)=p(x),\quad i=1,\cdots,C. 
\end{equation}
This indicates that $p(x|\mathcal{C}_i)=p(x)$ when Eq.~(\ref{eq22}) is minimized for this given image $x\in X$.
Thus, it can be easily obtained that,
\begin{equation}\label{eq23}
\begin{aligned}
&KL(p(x|\mathcal{C}_i)\parallel p(x))=\\
&\sum_{x\in X}p(x|\mathcal{C}_i)\log(\frac{p(x|\mathcal{C}_i)}{p(x)})=0, \quad \quad i=1,\cdots,C. 
\end{aligned}
\end{equation}\quad\quad\quad\quad\quad\quad \quad \quad\quad \quad \quad \quad \quad\quad\quad\quad\quad\quad\quad\quad\quad\quad\quad~ \quad$\blacksquare$

Based on the above result, we can readily infer the following corollary.

\textbf{Corollary 1.}
\textit{If the discriminator predicts the equal probability for any given image $x$ with respect to all camera classes, the data distributions of all cameras can be aligned as
\begin{equation*}
\begin{aligned}
KL(p(x|\mathcal{C}_1)\parallel p(x))=\cdots =KL(p(x|\mathcal{C}_C)\parallel p(x))=0.
\end{aligned}
\end{equation*}}

According to \textit{Proposition $1$}, a straightforward way may be to simply require an image from any camera classes to be equiprobably classified into all of the $C$ camera classes (i.e., ``all camera equiprobability'', ACE in short), which is similar to the method in the literature~\cite{DBLP:journals/corr/abs-1904-01308}. However, such an ACE scheme cannot produce satisfactory performance as expected, which will be sufficiently validated by our experiments. 

This can be explained as follows. First, the objective function in Eq.~(\ref{eq22}) is a complicated composite function of network parameters in the feature extractor $F$. As a result, the optimization can hardly achieve the global minimum. That's to say, the result of \textit{Proposition $1$} will not be exactly achieved in practice, and in turn this will not guarantee the expected reduction on the data distribution discrepancy. 

Second, a more subtle issue is that the result of \textit{Proposition $1$} (i.e., predicting the equal probability for all camera classes) is the goal set for the discriminator \textit{after the equilibrium is achieved through adversarial learning}. It shall not be simply used as the requirement to design the discriminator loss \textit{during adversarial learning}. As evidence, in order to make the final discriminator to predict equal probability (i.e., $1/2$) for both true and fake classes, the original GANs assign the class label ``1'' explicitly (rather than with a probability of $1/2$) to any sample from the fake class to design the discriminator loss used to train the generator~\cite{goodfellow2014generative}. Applying the same rule to our case, it means that we shall not follow the ACE scheme to design the discriminator loss in Eq.~(\ref{eq22}). 

To better reflect this situation, we develop a more precise scheme to achieve the multi-camera adversarial learning. In this new scheme, we require that with the learned feature representation, the discriminator shall classify an image into all camera classes with equal probability \textit{except the camera class which the image originally belongs to} (i.e., zero probability for this class).
This is where the name ``other camera equiprobability'' (OCE) comes from. This scheme can effectively avoid the effect of ``local camera assignment'', which is unfavorably done in the GRL-based scheme and make better efforts to align all cameras together, as demonstrated in the experiments. Furthermore, note that the proposed OCE-based scheme does not conflict with the result of \textit{Proposition $1$}, as analyzed above. As to be confirmed by Fig.~\ref{fig8} in Section~\ref{sec:EXP-FA}, the discriminator obtained by our scheme indeed has a clearer trend than the GRL-based scheme to predict the equal probability for all camera classes (i.e., \textit{Corollary $1$}).


 Formally, the proposed OCE scheme on an image $x^k$ in the $k$-th camera can be expressed as
\begin{equation}\label{eq13}
\begin{aligned}
\mathcal{L}_\mathrm{OCE}(x^k)=
 -\frac{1}{C-1}\sum_{\substack{i=1\\i\neq k}}^{C}\log (D(F(x^k),i)),
\end{aligned}
\end{equation}
where $D(F(x^k),i)$ denotes the predicted probability that $x^k$ belongs to the $i$-th camera class. In this way, the optimization for training the extractor feature $F$ (backbone network) is defined as
\begin{equation}\label{eq14}
\begin{aligned}
&\min_{F}\mathcal{L}_\mathrm{MCAL-F}(X, D)=\min_{F}\mathbb{E}_{x\sim X} \mathcal{L}_\mathrm{OCE}(x).
  \end{aligned}
\end{equation} 

Note that the traditional two-domain adversarial learning methods~\cite{DBLP:conf/icml/GaninL15,DBLP:conf/cvpr/TzengHSD17} are just a special case of this OCE-based scheme when there are exactly two camera classes on a dataset.

\begin{figure}
\centering
\includegraphics[width=8.5cm]{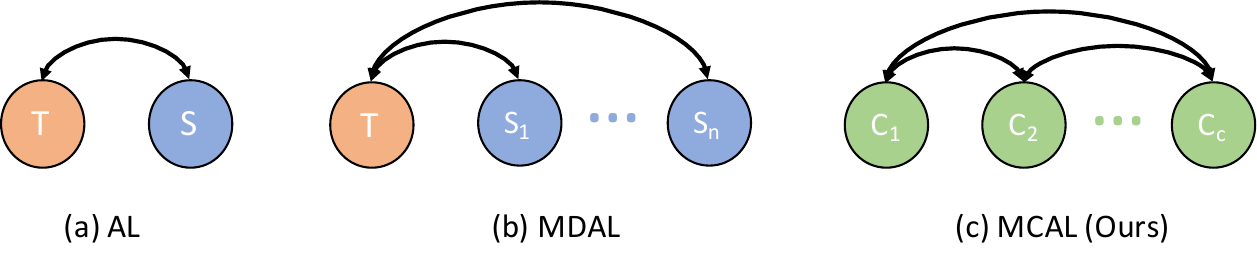}
\caption{Comparison among different adversarial schemes. (a) is adversarial learning (AL) in the conventional domain adaptation. (b) denotes the multi-domain adversarial learning (MDAL) for multi-source domain adaptation. (c) represents multi-camera adversarial learning (MCAL) for our camera alignment task. In this figure, T, S and C denote target domain, source domain and camera, respectively.}
\label{fig9}
\vspace*{-20pt}
\end{figure}

\textbf{Discussion.} a) \textit{Comparison among different adversarial learnings}. To show the difference between the proposed multi-camera adversarial leaning and other adversarial learning in the literature, we illustrate them in Fig.~\ref{fig9}. The conventional adversarial learning (AL) methods~\cite{DBLP:conf/icml/GaninL15,DBLP:journals/corr/abs-1803-09210,DBLP:conf/cvpr/TzengHSD17,wu2019few} conduct adversarial task between two domains (source and target domains or two different camera views). Multi-domain adversarial learning (MDAL)~\cite{zhao2018adversarial,xu2018deep,schoenauer-sebag2018multidomain,ghosh2018multi} deals with multi-source domain adaptation, which aligns one target domain and multiple source domains (i.e., ``one to many''). In contrast, the proposed multi-camera adversarial learning (MCAL) is specially designed for the camera alignment task, which conducts the camera adversarial task with each other (i.e., ``many to many'').

b) \textit{A caution on reducing distribution discrepancy.} 
As aforementioned, the OCE scheme can better reduce the distribution discrepancy among cameras than the GRL scheme, which is experimentally validated in Section~\ref{sec:EXP-FA}. 

Reducing data distribution discrepancy has been widely regarded as an effective means to mitigate the domain gap in computer vision. This also a key motivation of this work. Nevertheless, we would like to point out that this data discrepancy reduction approach may need to be used cautiously, for example, in the person Re-ID task. We observe that although the proposed OCE scheme is clearly better than the GRL scheme on distribution discrepancy reduction, its Re-ID performance could be less competitive than the latter. In other words, the relationship between discrepancy reduction and performance improvement may not be in a linear form. In some cases, single-minded pursuit of camera alignment could damage the intrinsic structure of the data distribution from each camera class, and this in turn adversely affects the final Re-ID performance. In Section~\ref{sec:EXP-EDC} and Section~\ref{sec:EXP-FA}, we will further discuss this issue by analyzing the results of two different adversarial schemes.
\subsection{Discrimination Learning with Intra-camera Labels}\label{SEC:LDI}
We utilize multi-camera adversarial learning to project all images of different cameras into a common feature space. Based on this shared space, we learn discriminative information by using the available within-camera label information. In this paper, we employ the triplet loss with the hard-sample-mining scheme~\cite{hermans2017defense}. It selects the hardest positive sample (i.e., the farthest positive sample) and the hardest negative sample (i.e., the closest negative sample) for an anchor to generate the triplet. This scheme can well optimize the embedding space such that data points with the same identity become closer to each other than those with different identities~\cite{hermans2017defense}. With the available within-camera label information, we only choose training data within each camera class to generate triplets. 
 In each batch, we randomly select $P$ persons and each person has $K$ images, i.e., $N=P\times K$. For the $c$-th camera, the triplet loss with hard sample mining can be described as
\begin{equation}\label{eq02}
\begin{aligned}
\min_{F}\mathcal{L}_{\mathrm{Triplet}}(X_{c},Y_{c})=\mathbb{E}_{(x,y)\sim (X_c,Y_c)}[m+l(x)]_{+}.
  \end{aligned}
\end{equation} For an anchor sample $x_{a}^{i}$ from the $i$-th person, 

\begin{equation}\label{eq03}
\begin{aligned}
l(x_{a}^{i})=\overbrace{\max_{p=1...K}d(F(x_{a}^{i}), F(x_{p}^{i}))}^{hardest~positive}-\overbrace{\min_{\substack{j=1...P\\ n=1,...K \\j\neq i}}d(F(x_{a}^{i}), F(x_{n}^{j}))}^{hardest~negative},
  \end{aligned}
\end{equation} and $m$ denotes the margin. $F(x_{a}^{i})$ is the feature of the sample $x_{a}^{i}$ and $d(\cdot,\cdot)$ is Euclidean distance of two feature vectors. 
\subsection{Optimization}\label{SEC:OPT}
The loss function of the proposed ACAN framework is finally expressed as 
\begin{equation}\label{eq10}
\begin{aligned}
&\min_{D}\mathcal{L}_\mathrm{MCAL-D}(X, Z, F),\\
&\min_{F}\left[ \mathcal{L}_\mathrm{Triplet}(X_c, Y_c)+\lambda\mathcal{L}_\mathrm{MCAL-F}(X,D)\right], c \in \left\{1,\dots, C\right\},
\end{aligned}
\end{equation} where $\lambda$ is the parameter to trade off the camera alignment task and the discrimination learning task.

Let $\theta _{F}$ and $\theta _{D}$  be the learnable parameters of feature extractor $F$ and discriminator $D$, respectively. For the GRL-based adversarial scheme, we can use the following stochastic updates as
\begin{equation}\label{eq06}
\begin{aligned}
&\theta _{D}=\theta _{D}-\mu \frac{\partial \mathcal{L}_\mathrm{MCAL-D}}{\partial \theta _{D}},\\
&\theta _{F}=\theta _{F}-\mu \left (\frac{\partial \mathcal{L}_{\mathrm{Triplet}}}{\theta _{F}} - \lambda \frac{\partial \mathcal{L}_\mathrm{MCAL-D}}{\partial \theta _{F}}\right ).
  \end{aligned}
\end{equation}  Since the GRL-based scheme employs the gradient reversal method, we directly use the reversal gradient of the discriminator to update the feature extractor $F$, as shown in Eq.~(\ref{eq06}).

For the OCE-based adversarial scheme, we can update the feature extractor $F$ and discriminator $D$ as
\begin{equation}\label{eq08}
\begin{aligned}
&\theta _{D}=\theta _{D}-\mu \frac{\partial \mathcal{L}_\mathrm{MCAL-D}}{\partial \theta_{D}},\\
&\theta _{F}=\theta _{F}-\mu \left (\frac{\partial \mathcal{L}_{\mathrm{Triplet}}}{\theta _{F}} + \lambda\frac{\partial \mathcal{L}_\mathrm{MCAL-F}}{\partial \theta _{F}}\right ),
  \end{aligned}
\end{equation} where $\mathcal{L}_\mathrm{MCAL-F}$ is defined in Eq.~(\ref{eq14}). In the optimizing process, since the discriminator needs to calculate two different losses (i.e., the cross-entropy loss for updating the parameters of discriminator and the OCE loss for updating the parameters of feature extractor), we employ the alternate way to update the discriminator $D$ and the feature extractor $F$, which is similar to the typical GANs methods~\cite{goodfellow2014generative,DBLP:conf/iccv/ZhuPIE17}. The pseudo-code of the whole optimization process is shown in Algorithm~\ref{alg1}.

\begin{algorithm}[ht]
\caption{\small{Adversarial Camera Alignment Network (ACAN)}}~\label{alg1}
\begin{algorithmic}[1]
\STATE {\bf procedure} 
ACAN (Training examples $\mathrm{X}$, labels $\mathrm{Y}_{Intra}$) \\
\STATE $\theta_{F} \leftarrow$ Initialize by ResNet-50 pre-trained on ImageNet.\\
\STATE $\theta_{D} \leftarrow$ Initialize by random number.\\
\FOR{epoch $\in [1,...,T]$}
\STATE \textcolor{blue}{// Fix feature extractor $\theta_{F}$, and update discriminator $\theta_{D}$.}\\
\STATE $\theta _{D}=\theta _{D}-\mu \frac{\partial \mathcal{L}_\mathrm{MCAL-D}}{\partial \theta _{D}}$\\
\STATE \textcolor{blue}{// Fix discriminator $\theta_{D}$, and update feature extractor $\theta_{F}$.}\\
\IF {using GRL scheme}
 \STATE  $\theta _{F}=\theta _{F}-\mu \left (\frac{\partial \mathcal{L}_{\mathrm{Triplet}}}{\theta _{F}} - \lambda \frac{\partial \mathcal{L}_\mathrm{MCAL-D}}{\partial \theta _{F}}\right )$. 
 \ELSE
 \STATE  $\theta _{F}=\theta _{F}-\mu \left (\frac{\partial \mathcal{L}_{\mathrm{Triplet}}}{\theta _{F}} + \lambda\frac{\partial \mathcal{L}_\mathrm{MCAL-F}}{\partial \theta _{F}}\right )$.
 \ENDIF
\ENDFOR
\STATE {\bf end procedure}
\end{algorithmic}
\end{algorithm}

\section{Experiments}\label{s-experiment}
In this part, we first introduce the experimental datasets and settings in Section~\ref{sec:EXP-DS}. Then, we compare the proposed method with the state-of-the-art unsupervised Re-ID methods and some methods with the UCC setting in Sections~\ref{sec:EXP-CUA} and~\ref{sec:EXP-SS}, respectively. To validate the effectiveness of various components in the proposed framework, we conduct ablation study in Section~\ref{sec:EXP-EDC}. Lastly, we further analyze the property of the proposed network in Section~\ref{sec:EXP-FA}.
\subsection{Datasets and Experiment Settings}\label{sec:EXP-DS}
We evaluate our approach on three large-scale image datasets: Market1501~\cite{DBLP:conf/iccv/ZhengSTWWT15}, DukeMTMC-reID~\cite{DBLP:conf/iccv/ZhengZY17}, and MSMT17~\cite{wei2018person}. 
 \textbf{Market1501} contains 1,501 persons with 32,668 images from six cameras. Among them, $12,936$ images of $751$ identities are used as a training set. For evaluation, there are $3,368$ and $19,732$ images in the query set and the gallery set, respectively. \textbf{DukeMTMC-reID} has $1,404$ persons from eight cameras, with $16,522$ training images, $2,228$ queries, and $17,661$ gallery images.
 \textbf{MSMT17}\footnote{On MSMT17, the second camera includes $15$ identities with $193$ images. In each training batch, since 32 identities are chosen from each camera, we do not select data from the camera in all experiments. Therefore, we use the data of $14$ cameras to train our model on this dataset.}
 is collected from a 15-camera network deployed on campus. The training set contains $32,621$ images of $1,041$ identities. For evaluation, $11,659$ and $82,161$ images are used as query and gallery images, respectively. For all datasets, we employ CMC (i.e., Cumulative Match Characteristic) accuracy and mAP (i.e., mean Average Precision) for Re-ID evaluation~\cite{DBLP:conf/iccv/ZhengSTWWT15}. On Market1501, there are single- and multi-query evaluation protocols. We use the more challenging single-query protocol in our experiments.

In addition, we also evaluate the proposed method on two large-scale video datasets including MARS~\cite{DBLP:conf/eccv/ZhengBSWSWT16} and DukeMTMC-SI-Tracklet~\cite{li2019unsupervised}. \textbf{MARS} has a total of $20,478$ tracklets of $1,261$ persons captured from a 6-camera network on a university campus. All the tracklets were automatically generated by the DPM detector~\cite{felzenszwalb2009object} and the GMMCP tracker~\cite{dehghan2015gmmcp}. This dataset splits $626$ and $635$ identities into training and testing sets, respectively. \textbf{DukeMTMC-SI-Tracklet} is from DukeMTMC. It consists of $19,135$ person tracklets and $1,788$ persons from $8$ cameras. In this dataset, $702$ and $1,086$ identities are used to train and evaluate models, respectively. On the video datasets, we also employ CMC accuracy and mAP for Re-ID evaluation~\cite{DBLP:conf/iccv/ZhengSTWWT15}.

For training the multi-camera adversarial learning task, we randomly select the same number (i.e., $\left \lfloor \frac{64}{C} \right \rfloor$, where $C$ is the number of cameras and $\left \lfloor \cdot  \right \rfloor$ denotes the round down operation) of images per camera in a batch. To generate triplets, we set $P$ (i.e., number of persons) and $K$ (i.e., number of images per person) as $32$ and $4$ in each training batch, which is the same with the literature~\cite{hermans2017defense}. The margin of triplet loss, $m$, is $0.3$ according to the literature~\cite{hermans2017defense}. $\lambda$ in Eq. (\ref{eq10}) is set as $1$. We use a fully connected layer to implement the discriminator $D$, whose dimension is set as $128$.
 The initial learning rates of the fine-tuned parameters (those in the pre-trained ResNet-50 on ImageNet~\cite{DBLP:conf/cvpr/DengDSLL009}) and the new parameters (those in the newly added layers) are $0.1$ and $0.01$, respectively. The proposed model is trained with the SGD optimizer in a total of $300$ epochs. When the number of epochs reaches $100$ and $200$, we decrease the learning rates by a factor of $0.1$. The size of the input image is $256 \times 128$. Particularly, all experiments on all datasets utilize the same experimental settings.

\subsection{Comparison with Unsupervised Methods}\label{sec:EXP-CUA}
\renewcommand{\cmidrulesep}{0mm} 
\setlength{\aboverulesep}{0mm} 
\setlength{\belowrulesep}{0mm} 
\setlength{\abovetopsep}{0cm}  
\setlength{\belowbottomsep}{0cm}
\begin{table*}[htbp]
  \centering
  \caption{Comparison with the state-of-the-art unsupervised methods on three image datasets including Market1501, DukeMTMC-reID and MSMT17. ``-'' denotes that the result is not provided. We report mAP and the Rank-1, 5, 10 accuracies of CMC. The best performance is shown in \textbf{bold}.}
    \begin{tabular}{|c|cccc|cccc|cccc|}
    \toprule
    \midrule
    \multirow{2}[1]{*}{Method} & \multicolumn{4}{c|}{Market1501} & \multicolumn{4}{c|}{DukeMTMC-reID} & \multicolumn{4}{c|}{MSMT17} \\
\cmidrule{2-13}          & mAP   & Rank-1 & Rank-5 & Rank-10 & mAP   & Rank-1 & Rank-5 & Rank-10 & mAP   & Rank-1 & Rank-5 & Rank-10 \\
    \midrule
    LOMO~\cite{liao2015person}  & 8.0   & 27.2  & 41.6  & 49.1  & 4.8   & 12.3  & 21.3  & 26.6  & -    & -    & -    & - \\
    BoW~\cite{DBLP:conf/iccv/ZhengSTWWT15}   & 14.8  & 35.8  & 52.4  & 60.3  & 8.3   & 17.1  & 28.8  & 34.9  & -    & -    & -    & - \\
    UJSDL~\cite{qi2018unsupervised} & -    & 50.9  & -    & -    &    -   & 32.2  & -    & -    & -    & -    & -    & - \\
    UMDL~\cite{DBLP:conf/cvpr/PengXWPGHT16}  & 12.4  & 34.5  & -    & -    & 7.3   & 18.5  & -    & -    & -    & -    & -    & - \\
    CAMEL~\cite{DBLP:conf/iccv/YuWZ17} & 26.3  & 54.5  & -    & -    &   -    &   -    & -    & -    & -    & -    & -    & - \\
    \midrule
    PUL~\cite{fan2018unsupervised}   & 20.5  & 45.5  & 60.7  & 66.7  & 16.4  & 30.0  & 43.4  & 48.5  & -    & -    & -    & - \\
    Tfusion~\cite{lv2018unsupervised} & -    & 60.8  & -    & -    & -    & -    & -    & -    & -    & -    & -    & - \\
    TJ-AIDL~\cite{wang2018transferable}  & 26.5  & 58.2  & 74.8  & 81.1  & 23.0  & 44.3  & 59.6  & 65.0  & -    & -    & -    & - \\
    \midrule
    MMFA~\cite{DBLP:conf/bmvc/LinLLK18}  & 27.4  & 56.7  & 75.0  & 81.8  & 24.7  & 45.3  & 59.8  & 66.3  & -    & -    & -    & - \\
    CAT~\cite{DBLP:journals/corr/abs-1904-01308}   & 27.8  & 57.8  & -    & -    & 28.7  & 50.9  & -    & -    & -    & -    & -    & - \\
    CAL-CCE~\cite{qi2019novel} & 34.5  & 64.3  & -    & -    & 36.7  & 55.4  & -    & -    & -    & -    & -    & - \\
    \midrule
    PTGAN~\cite{wei2018person} & -    & 38.6  & -    & -    & -    & 27.2  & -    & -    & 3.3   & 11.8  & -    & 27.4 \\
    SPGAN~\cite{deng2018image} & 22.8  & 51.5  & 70.1  & 76.8  & 22.3  & 41.1  & 56.6  & 63.0  & -    & -    & -    & - \\
    SPGAN+LMP~\cite{deng2018image}  & 26.7  & 57.7  & 75.8  & 82.4  & 26.2  & 46.4  & 62.3  & 68.0  & -    & -    & -    & - \\
    HHL~\cite{zhong2018generalizing}   & 31.4  & 62.2  & 78.8  & 84.0  & 27.2  & 46.9  & 61.0  & 66.7  & -    & -    & -    & - \\
    UTA~\cite{tian2019imitating}   & 40.1  & 72.4  & 87.4  & 91.4  & 31.8  & 55.6  & 68.3  & 72.4  & -    & -    & -    & - \\
    ECN~\cite{zhong2019invariance}   & 43.0  & \textbf{75.1} & \textbf{87.6}  & 91.6  & 40.4  & 63.3  & 75.8  & 80.4  & 10.2  & 30.2  & 41.5  & 46.8 \\
    CSGLP~\cite{DBLP:journals/tifs/RenLGZL20}   & 31.5  & 61.2  & 77.5  & 83.2  & 27.1  & 47.8  & 62.3  & 68.3  & -    & -    & -    & - \\
    \midrule
    ACAN-GRL  (ours) & \textbf{50.6} & 73.3  & \textbf{87.6} & \textbf{91.8} & \textbf{46.6} & 65.1  & 80.6  & 85.1  & 11.2  & 27.1  & 40.9  & 47.3 \\
    ACAN-OCE  (ours) & 47.7  & 72.2  & 86.3  & 90.4  & 45.1  & \textbf{67.6} & \textbf{81.2} & \textbf{85.2} & \textbf{12.6} & \textbf{33.0} & \textbf{48.0} & \textbf{54.7} \\
    \bottomrule
    \end{tabular}%
  \label{tab01}%
  \vspace*{-20pt}
\end{table*}%

We compare our method with the state-of-the-art unsupervised image-based person Re-ID methods. 
Among them, there are five non-deep-learning-based methods (LOMO~\cite{liao2015person}, BoW~\cite{DBLP:conf/iccv/ZhengSTWWT15}, UJSDL~\cite{qi2018unsupervised}, UMDL~\cite{DBLP:conf/cvpr/PengXWPGHT16} and CAMEL~\cite{DBLP:conf/iccv/YuWZ17}), and multiple deep-learning-based methods. The latter includes three recent pseudo-label-generation-based methods (PUL~\cite{fan2018unsupervised}, TFusion~\cite{lv2018unsupervised} and TJ-AIDL~\cite{wang2018transferable}), three distribution-alignment-based methods (MMFA~\cite{DBLP:conf/bmvc/LinLLK18}, CAT~\cite{DBLP:journals/corr/abs-1904-01308} and CAL-CCE~\cite{qi2019novel}) and five recent image-generation-based approaches (PTGAN~\cite{wei2018person}, SPGAN~\cite{deng2018image}, HHL~\cite{zhong2018generalizing},  UTA~\cite{tian2019imitating}, ECN~\cite{zhong2019invariance} and CSGLP~\cite{DBLP:journals/tifs/RenLGZL20}). Particularly, although there is no label information for the target domain in these unsupervised methods, most of them utilize the labeled source domains or generated images by the GAN-based models. Differently, our method does not utilize such information. The experimental results on Market1501, DukeMTMC-reID and MSMT17 are reported in Table~\ref{tab01}. As seen, the results consistently show the superiority of our proposed method over the aforementioned methods. For example, the proposed method significantly outperforms the recent pseudo-label-generation-based methods, such as PUL and CAMEL. This is contributed to the available intra-camera label information and the multi-camera adversarial learning method which can effectively align the distributions from different cameras. Particularly, compared with the recent ECN~\cite{zhong2018generalizing}, the state-of-the-art by utilizing both the labeled source domain and generated images from the GAN-based model, our ACAN-OCE still gains $4.7\%$ ($47.7$ vs. $43.0$), $4.7\%$ ($45.1$ vs. $40.4$) and $2.4\%$ ($12.6$ vs. $10.2$) in mAP on Market1501, DukeMTMC-reID and MSMT17, respectively. In particular, the proposed method does not use any other extra data, such as labeled source domains or generated images by the GAN-based methods~\cite{wei2018person,deng2018image}.

Moreover, we also compare our method with the state-of-the-art unsupervised video-based Re-ID approaches on MARS and DukeMTMC-SI-Tracklet, which include
GRDL~\cite{kodirov2016person}, UnKISS~\cite{khan2016unsupervised}, RACE~\cite{ye2018robust}, Stepwise~\cite{liu2017stepwise}, DGM+MLAPG~\cite{ye2017dynamic}, DGM+IDE~\cite{ye2017dynamic}, DAL~\cite{DBLP:conf/bmvc/ChenZG18}, TAUDL~\cite{Li_2018_ECCV} and UTAL~\cite{li2019unsupervised}. Among them, except for GRDL and UnKISS, all other methods are based on deep features. The results are reported in Table~\ref{tab02}. Compared with the recent UTAL, which has shown the superiority in the unsupervised video-based Re-ID task, our method (ACAN-GRL or ACAN-OCE) achieves a significant improvement in mAP and CMC accuracy. In particular, ACAN-GRL outperforms UTAL by $13.9\%$ ($49.1$ vs. $35.2$) and $9.3\%$ ($59.2$ vs. $49.9$) in mAP and Rank-1 accuracy on MARS, respectively. This further validates the effectiveness of our proposed methods.

\begin{table}[htbp]
  \centering
  \caption{Comparison with the state-of-the-art unsupervised methods on two video datasets including MARS and DukeMTMC-SI-Tracklet (Duke-T).}
    \begin{tabular}{|c|c|cccc|}
    \toprule
    \midrule
          & Method & mAP   & Rank-1 & Rank-5 & Rank-20 \\
          \midrule
    \multirow{11}[1]{*}{\begin{sideways}MARS\end{sideways}} & GRDL~\cite{kodirov2016person}  & 9.6   & 19.3  & 33.2  & 46.5 \\
          & UnKISS~\cite{khan2016unsupervised} & 10.6  & 22.3  & 37.4  & 53.6 \\
          & RACE~\cite{ye2018robust}  & 24.5  & 43.2  & 57.1  & 67.6 \\
          & Stepwise~\cite{liu2017stepwise} & 10.5  & 23.6  & 35.8  & 44.9 \\
          & DGM+MLAPG~\cite{ye2017dynamic} & 11.8  & 24.6  & 42.6  & 57.2 \\
          & DGM+IDE~\cite{ye2017dynamic} & 21.3  & 36.8  & 54.0  & 68.5 \\
          & DAL~\cite{DBLP:conf/bmvc/ChenZG18}   & 21.4  & 46.8  & 63.9  & 77.5 \\
          & TAUDL~\cite{Li_2018_ECCV} & 29.1  & 43.8  & 59.9  & 72.8 \\
          & UTAL~\cite{li2019unsupervised}  & 35.2  & 49.9  & 66.4  & 77.8 \\
\cmidrule{2-6}          & ACAN-GRL (ours) & \textbf{49.1} & \textbf{59.2} & \textbf{77.1} & \textbf{86.7} \\
          & ACAN-OCE (ours) & 47.5  & 57.7  & 75.1  & 84.0 \\
    \midrule
    \midrule
    \multirow{4}[1]{*}{\begin{sideways}Duke-T\end{sideways}} & TAUDL~\cite{Li_2018_ECCV} & 20.8  & 26.1  & 42.0  & 57.2 \\
          & UTAL~\cite{li2019unsupervised}  & 36.6  & 43.8  & 62.8  & 76.5 \\
\cmidrule{2-6}          & ACAN-GRL (ours) & \textbf{43.0} & \textbf{52.0} & \textbf{71.0} & \textbf{82.0} \\
          & ACAN-OCE (ours) & 40.3  & 50.4  & 68.0  & 81.2 \\
    \bottomrule
    \end{tabular}%
  \label{tab02}%
  \vspace*{-20pt}
\end{table}%

\subsection{Comparison with Methods in UCC Setting} \label{sec:EXP-SS}
In this section, we compare our method with the unsupervised video-based person Re-ID methods working in the UCC setting (i.e., given the intra-camera label information). Both TAUDL~\cite{Li_2018_ECCV} and UTAL~\cite{li2019unsupervised} consist of intra-camera tracklet discrimination learning and cross-camera tracklet association learning. In the UCC setting, they construct the connection of different cameras by self-discovering the cross-camera positive matching pairs. Different from both TAUDL~\cite{Li_2018_ECCV} and UTAL~\cite{li2019unsupervised}, we address the cross-camera problem from the data-distribution alignment perspective. We show the experimental results in Tables~\ref{tab03} and~\ref{tab04}. As seen, on all image datasets (Table~\ref{tab03}), the proposed method has a better performance with TAUDL and UTAL. For example, the Rank-1 accuracy of ACAN-OCE gains $3.0\%$ ($72.2$ vs. $69.2$), $5.3\%$ ($67.6$ vs. $62.3$) and $1.6\%$ ($33.0$ vs. $31.4$) over UTAL on Market1501, DukeMTMC-reID and MSMT17, respectively. Besides, on video datasets, the proposed method can also achieve competitive performance. Although the results are slightly worse when compared with UTAL on MARS, the proposed method still obtains a large improvement on DukeMTMC-SI-Tracklet. 
As seen, on DukeMTMC-SI-Tracklet, ACAN-GRL can improve $4.0\%$ ($43.0$ vs. $39.0$) and $5.6\%$ ($52.0$ vs. $46.4$) over UTAL in mAP and Rank-1 accuracy, respectively.
Particularly, we also further demonstrate the effectiveness of multi-camera adversarial learning on all datasets in Section~\ref{sec:EXP-EDC}.

\newcommand{\PreserveBackslash}[1]{\let\temp=\\#1\let\\=\temp}
\newcolumntype{C}[1]{>{\PreserveBackslash\centering}p{#1}}
\newcolumntype{R}[1]{>{\PreserveBackslash\raggedleft}p{#1}}
\newcolumntype{L}[1]{>{\PreserveBackslash\raggedright}p{#1}}
\begin{table}[htbp]
  \centering
  \caption{Comparison with TAUDL and UTAL in the UCC setting on Market1501, DukeMTMC-reID (Duke) and MSMT17. The best performance is shown in \textbf{bold}.}
    \begin{tabular}{|C{2.2cm}|C{0.3cm}C{0.9cm}|C{0.3cm}C{0.9cm}|C{0.3cm}C{0.9cm}|}
    \toprule
    \midrule
    \multirow{2}[1]{*}{Method} & \multicolumn{2}{c|}{Market1501} & \multicolumn{2}{c|}{Duke} & \multicolumn{2}{c|}{MSMT17} \\
\cmidrule{2-7}          & mAP   & Rank-1 & mAP   & Rank-1 & mAP & Rank-1 \\
    \midrule
    TAUDL~\cite{Li_2018_ECCV} & 41.2  & 63.7  & 43.5  & 61.7  & 12.5 & 28.4 \\
    UTAL~\cite{li2019unsupervised}  & 46.2  & 69.2  & 44.6  & 62.3  &  \textbf{13.1} & 31.4 \\
    \midrule
    ACAN-GRL (ours) & \textbf{50.6} & \textbf{73.3} & \textbf{46.6} & 65.1  & 11.2 & 27.1 \\
    ACAN-OCE (ours) & 47.7  & 72.2  & 45.1  & \textbf{67.6} & 12.6 & \textbf{33.0} \\
    \bottomrule
    \end{tabular}%
  \label{tab03}%
  \vspace*{-20pt}
\end{table}%

\begin{table}[htbp]
  \centering
  \caption{Comparison with UTAL in the UCC setting on MARS and DukeMTMC-SI-Tracklet (Duke-T).}
    \begin{tabular}{|c|cc|cc|}
    \toprule
    \midrule
    \multirow{2}[1]{*}{Method } & \multicolumn{2}{c|}{MARS} & \multicolumn{2}{c|}{Duke-T} \\
\cmidrule{2-5}          & mAP   & Rank-1 & mAP   & Rank-1 \\
    \midrule
    UTAL~\cite{li2019unsupervised}  & \textbf{51.7} & \textbf{59.5} & 39.0  & 46.4 \\
    \midrule
    ACAN-GRL (ours) & 49.1  & 59.2  & \textbf{43.0} & \textbf{52.0} \\
    ACAN-OCE (ours) & 47.5  & 57.7  & 40.3  & 50.4 \\
    \bottomrule
    \end{tabular}%
  \label{tab04}%
  \vspace*{-10pt}
\end{table}%

\subsection{Effectiveness of Different Components in ACAN}\label{sec:EXP-EDC}
To sufficiently validate the efficacy of different components in the proposed adversarial camera alignment network, we conduct experiments on five large-scale datasets. The experimental results are reported in Table \ref{tab05}. Firstly, on all datasets, using multi-camera adversarial learning to align all cameras can significantly improve the performance of the model conducting the only intra-camera discrimination task (i.e., ``only $\mathcal{L}_{\mathrm{Triplet}}$'' in Table \ref{tab05}). For example, ACAN-GRL can improve $14.8\%$ ($49.1$ vs. $34.3$) and $12.3\%$ ($43.0$ vs. $30.7$) in mAP on MARS and DukeMTMC-SI-Tracklet. Also, the Rank-1 accuracy can be improved by $13.7\%$ ($59.2$ vs. $45.5$) and $13.6\%$ ($52.0$ vs. $38.4$). This demonstrates the effectiveness of the proposed MCAL. Secondly, compared with results of PUL~\cite{fan2018unsupervised}, TFusion~\cite{lv2018unsupervised} and TJ-AIDL~\cite{wang2018transferable} in Table~\ref{tab01}, which are based on pseudo-label-generation, ``only $\mathcal{L}_{\mathrm{Triplet}}$'' still has the competitive performance. For example, ``only $\mathcal{L}_{\mathrm{Triplet}}$'' improves $7.9\%$ ($34.4$ vs. $26.5$) and $18.7\%$ ($41.7$ vs. $23.0$) over TJ-AIDL in mAP on Market1501 and DukeMTMC-reID, respectively. This shows using intra-camera labels indeed contributes to our task. Lastly, the OCE-based scheme is better than the GRL-based scheme on MSMT17 collected from 15 cameras, due to the limitation of GRL (i.e., maximize Eq.~(\ref{eq12}) by freely assigning camera labels), as analyzed in Section~\ref{SEC:AC}. However, for the datasets with few cameras, the OCE scheme may slightly damage the original intrinsic structure of data distribution due to the strict label-assignment strategy. We will further discuss the OCE and GRL schemes in Section~\ref{sec:EXP-FA}.
\begin{table}[htbp]
  \centering
  \caption{Evaluation of different components of the proposed ACAN on three image datasets (i.e., Market1501, DukeMTMC-reID and MSMT17) and two video datasets (i.e., DukeMTMC-SI-Tracklet (Duke-T) and MARS). The best performance is shown in \textbf{bold}.}
    \begin{tabular}{|c|c|ccc|}
    \toprule
      \midrule
     Dataset & Method & mAP   & Rank-1 & \multicolumn{1}{l|}{Rank-5} \\
    \midrule
    \multirow{3}[1]{*}{Market1501} & Only $\mathcal{L}_{\mathrm{Triplet}}$ & 34.4  & 58.1  & 72.5 \\
\cmidrule{2-5}          & ACAN-GRL (ours) & \textbf{50.6} & \textbf{73.3} & \textbf{87.6} \\
          & ACAN-OCE (ours) & 47.7  & 72.2  & 86.3 \\
    \midrule
    \midrule
    \multirow{3}[1]{*}{Duke} & Only $\mathcal{L}_{\mathrm{Triplet}}$ & 41.7  & 60.1  & 75.1 \\
\cmidrule{2-5}          & ACAN-GRL (ours) & \textbf{46.6} & 65.1  & 80.6 \\
          & ACAN-OCE (ours) & 45.1  & \textbf{67.6} & \textbf{81.2} \\
    \midrule
    \midrule
    \multirow{3}[1]{*}{MSMT17} & Only $\mathcal{L}_{\mathrm{Triplet}}$ & 10.0  & 24.8  & 38.1 \\
\cmidrule{2-5}          & ACAN-GRL (ours) & 11.2  & 27.1  & 40.9 \\
          & ACAN-OCE (ours) & \textbf{12.6} & \textbf{33.0} & \textbf{48.0} \\
    \midrule
    \midrule
    \multirow{3}[1]{*}{MARS} & Only $\mathcal{L}_{\mathrm{Triplet}}$ & 34.3  & 45.5  & 61.0 \\
\cmidrule{2-5}          & ACAN-GRL (ours) & \textbf{49.1} & \textbf{59.2} & \textbf{77.1} \\
          & ACAN-OCE (ours) & 47.5  & 57.7  & 75.1  \\
    \midrule
    \midrule
    \multirow{3}[1]{*}{Duke-T} & Only $\mathcal{L}_{\mathrm{Triplet}}$ & 30.7  & 38.4  & 55.5 \\
\cmidrule{2-5}          & ACAN-GRL (ours) & \textbf{43.0} & \textbf{52.0} & \textbf{71.0} \\
          & ACAN-OCE (ours) & 40.3  & 50.4  & 68.0 \\
    \bottomrule
    \end{tabular}%
  \label{tab05}%
  \vspace*{-20pt}
\end{table}%

\subsection{Further Analysis}\label{sec:EXP-FA}

\textbf{Algorithm Convergence.}
To investigate the convergence of our algorithm, we record the mAP and Rank-1 accuracy of ACAN-GRL and ACAN-OCE during training on a validated set of DukeMTMC-reID in Fig.~\ref{fig6}. As seen, our methods can almost converge after 200 epochs. 

\textbf{Running Time.}
For the feature extracting stage, each image needs to take 2.73 ms on GPU (NVIDIA 1080ti), which means each second can conduct about 366 person images. Besides, on the Market1501 dataset, which includes 3369 query images and 19733 gallery images, the matching process of all images merely needs 0.98 s. Therefore, our method can be deployed in a real-world scenario. 

\begin{figure}[htbp]
\centering
\begin{minipage}[t]{0.22\textwidth}
\centering
\includegraphics[width=4cm]{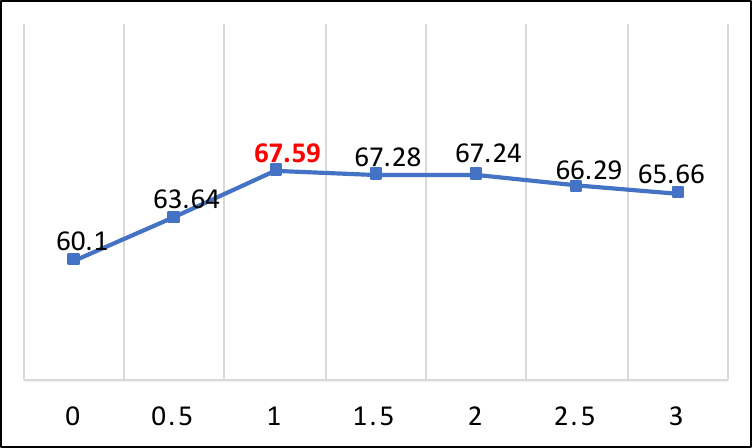}
\caption{Evaluation for different $\lambda$ in Eq.~(\ref{eq10}) on DukeMTMC-reID. Note that this result is reported on the test set.}
\label{fig4}
\end{minipage}
\begin{minipage}[t]{0.22\textwidth}
\centering
\includegraphics[width=4cm]{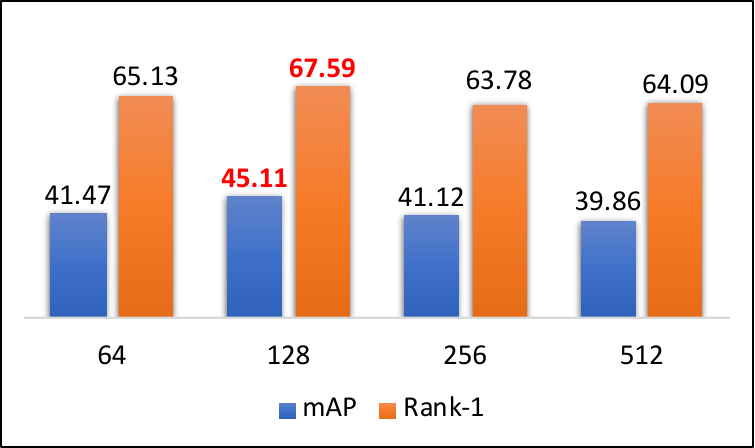}
\caption{Experimental results of different dimensional discriminators (i.e., different dimensional FC layers) on DukeMTMC-reID.}
\label{fig5}
\end{minipage}
\end{figure}

\begin{figure}
\centering
\subfigure[ACAN-GRL]{
\includegraphics[width=4cm]{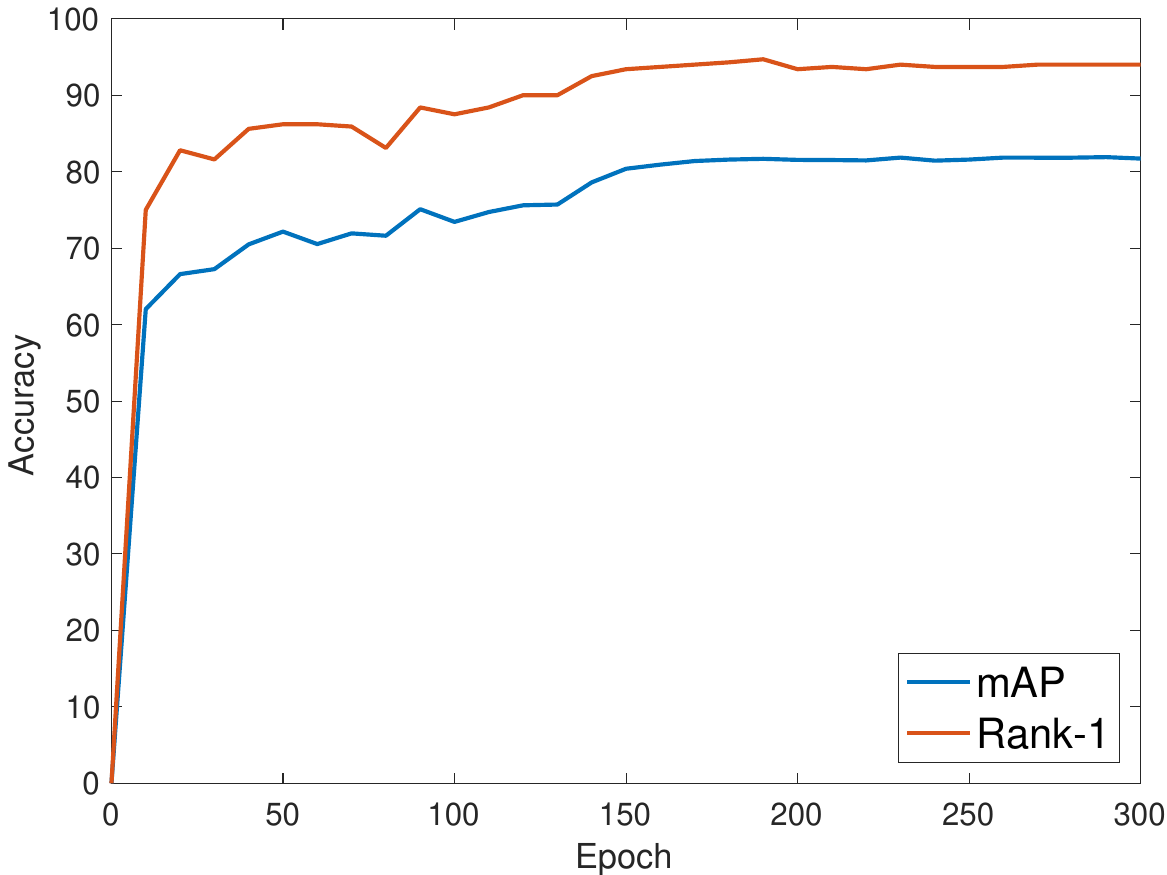}
}
\subfigure[ACAN-OCE]{
\includegraphics[width=4cm]{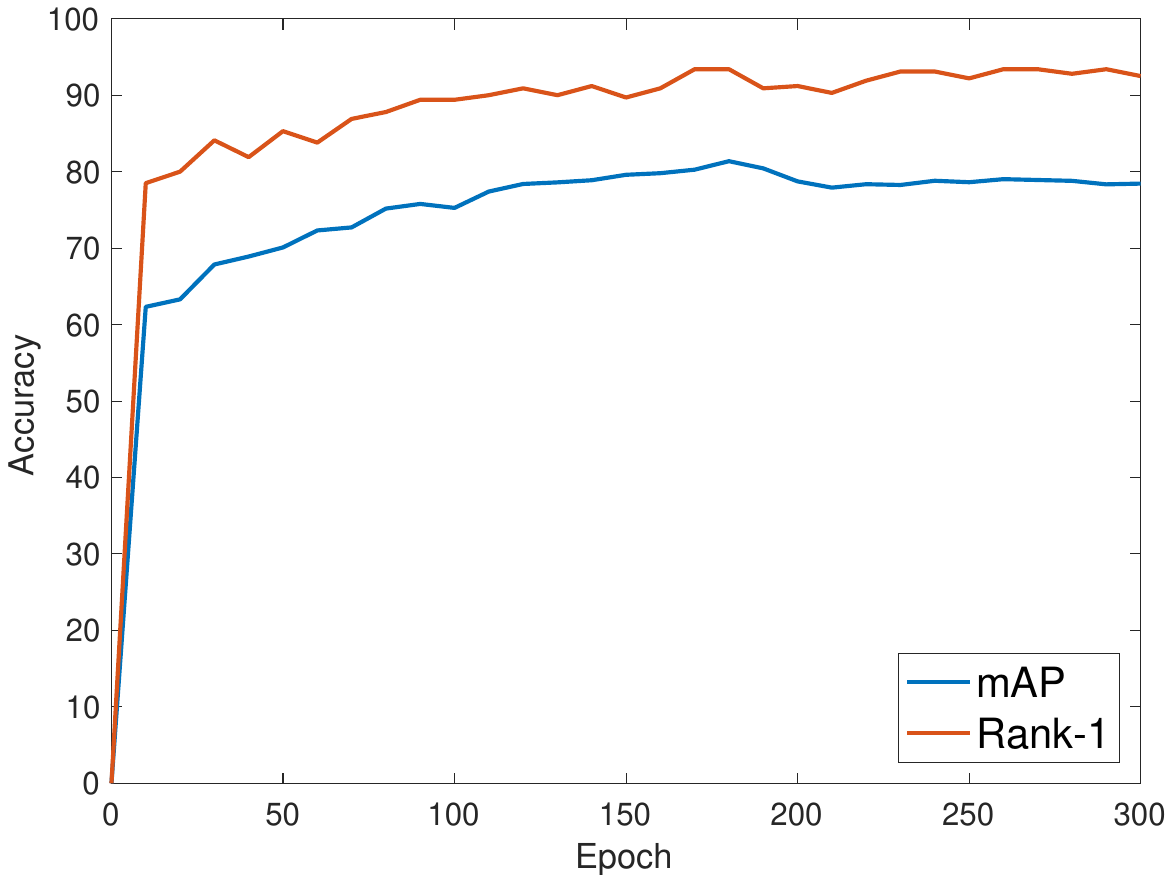}
}
\caption{Convergence curves of ACAN-GRL and ACAN-OCE on DukeMTMC-reID.}
\label{fig6}
\vspace*{-20pt}
\end{figure}

\textbf{Parameter Sensitivity.}
 To study the sensitivity of $\lambda$ in Eq.~(\ref{eq10}), which is the parameter to trade off the camera-alignment task and the intra-camera discrimination task, we sample the values in $\left\{0.0, 0.5, 1.0, 1.5, 2.0, 2.5, 3.0\right\}$, and perform the experiments by ACAN-OCE on DukeMTMC-reID. All the results are shown in Fig~\ref{fig4}, and we find that the Rank-1 accuracy first increases and then decreases. Finally, we set $\lambda = 1$ in all experiments for all datasets. Also, to analyze the sensitivity of the dimension of the discriminator in our network, we sample the values in $\left\{64, 128, 256, 512\right\}$, and perform the experiments by ACAN-OCE on DukeMTMC-reID. Fig.~\ref{fig5} shows the experimental results. As seen, we can obtain good performance when setting the dimension as 128. Consequently, we utilize a 128-dimensional discriminator (i.e., FC layer) in all experiments for all datasets.

\textbf{OCE VS. ACE.}
As analyzed in Section~\ref{SEC:AC-2}, we prefer the OCE-based scheme rather than the ACE-based scheme. In this part, we conduct a comparison between ACE and OCE in Table~\ref{tab06}. As shown, the OCE-based scheme outperforms the ACE-based scheme on all datasets. This is consistent with our previous analysis. For example, compared with the ACE scheme, the OCE scheme increases $7.9\%$ ($47.7$ vs. $39.8$) and $8.2\%$ ($72.2$ vs. $64.0$) in mAP and Rank-1 on Market1501, respectively. It indicates that during reducing camera-level discrepancy because i) it is difficult to achieve the ideal minimization in Eq.~(\ref{eq22}); ii) Since the ACE-based scheme does not follow the rule of the typical GANs~\cite{goodfellow2014generative} to design the discriminator loss, it cannot effectively achieve the multi-camera adversarial task. The concrete analysis is given in Section~\ref{SEC:AC-2}.

\begin{figure}
\centering

\subfigure[Duke (Baseline)]{
\includegraphics[width=4cm]{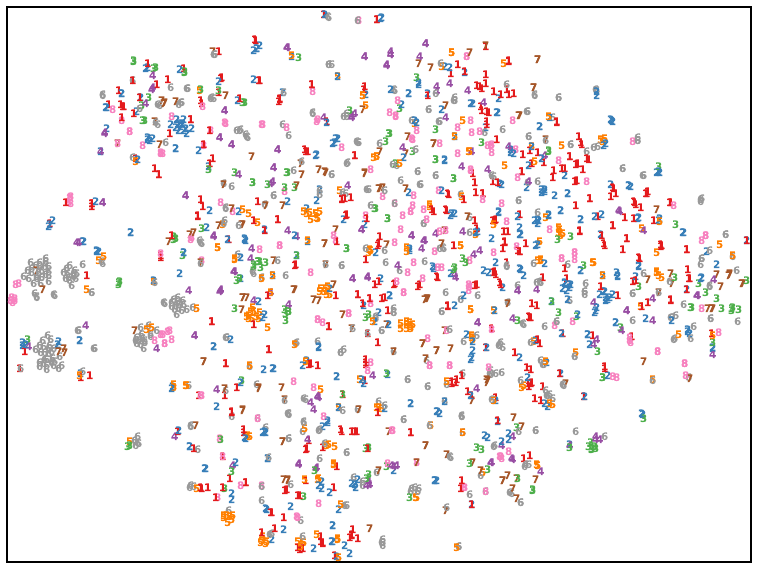}
}
\subfigure[MSMT (Baseline)]{
\includegraphics[width=4cm]{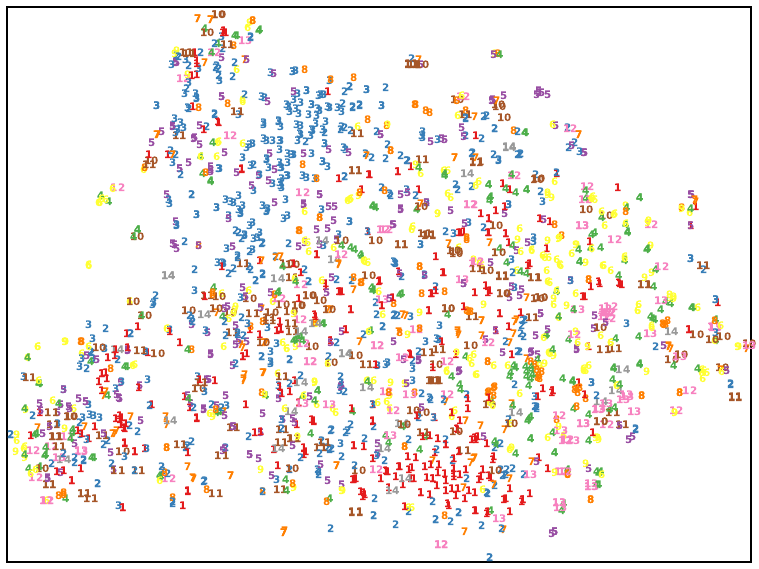}
}
\subfigure[Duke (ACAN-GRL)]{
\includegraphics[width=4cm]{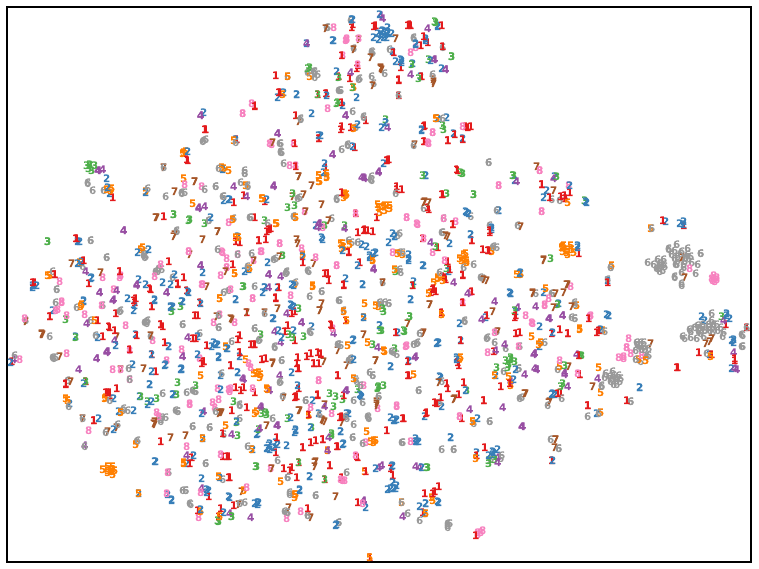}
}
\subfigure[MSMT (ACAN-GRL)]{
\includegraphics[width=4cm]{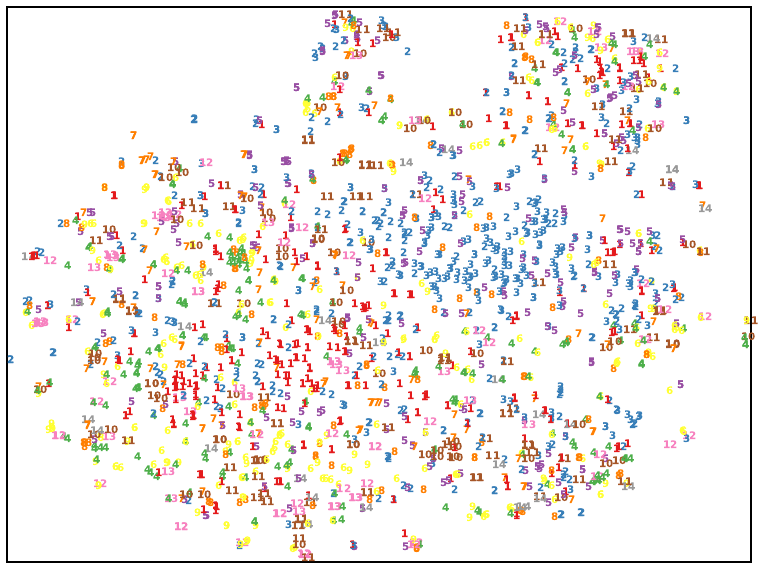}
}
\subfigure[Duke (ACAN-OCE)]{
\includegraphics[width=4cm]{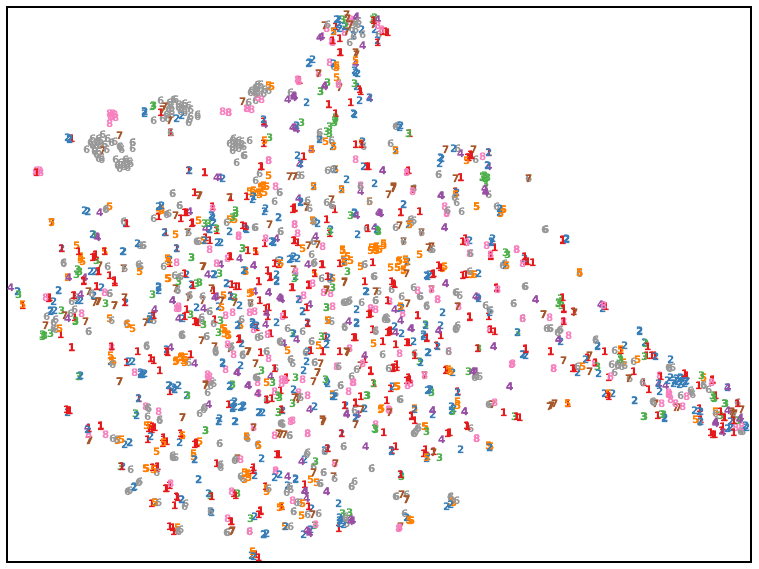}
}
\subfigure[MSMT (ACAN-OCE)]{
\includegraphics[width=4cm]{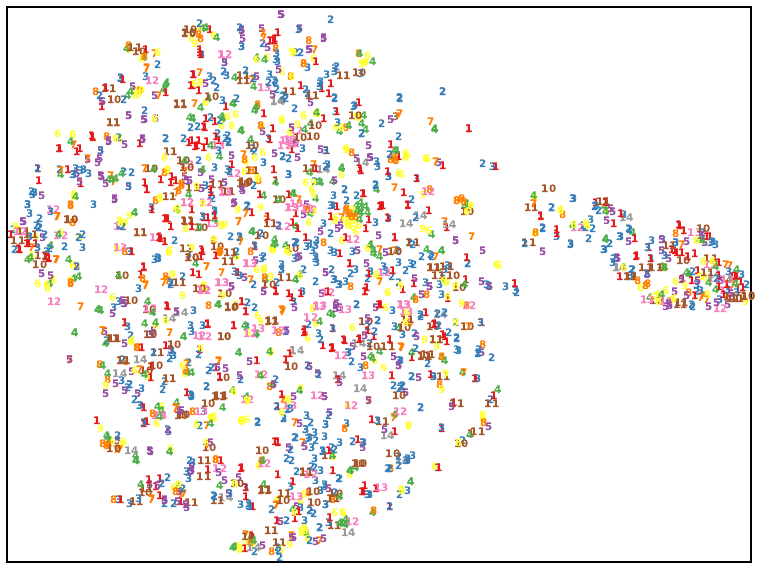}
}
\caption{Visualization of the distributions of two datasets via t-SNE~\cite{maaten2008visualizing}. The features of images are extracted by Baseline ((a) and (b)), ACAN-GRL ((c) and (d)) and ACAN-OCE ((e) and (f)). Different colors denote the images from different cameras. In detail, (a) (c) and (e) are eight different cameras on DukeMTMC-reID; (b) (d) and (f) are fourteen different cameras on MSMT17. Note that Baseline denotes ``Only $\mathcal{L}_{\mathrm{Triplet}}$'' in Eq.~(\ref{eq10}). Particularly, the visualization corresponds to Table~\ref{tab07}. Best viewed by vertical contrast.}
\label{fig7}
\vspace*{-20pt}
\end{figure}

\textbf{Distribution Visualization.}
We examine the inter-camera (cross-camera) discrepancy to validate the effectiveness of GRL and OCE.  In this experiment, to measure the inter-camera discrepancy, we use $d_{\mathrm{inter-camera}}=\frac{1}{C}\sum_{c=1}^{C}\left \| \overline{X}_{c}-\overline{X} \right \|_{2}$, where $\overline{X}_{c}$   is the sample mean of the $c$-th camera class and $\overline{X}$ denotes the mean of all samples. $C$ is the total number of cameras. These distances are calculated in Table~\ref{tab07}, where Baseline denotes ``Only $\mathcal{L}_{\mathrm{Triplet}}$'' in Eq.~(\ref{eq10}).
Firstly, for the inter-camera discrepancy, both the GRL-based and OCE-based schemes achieve smaller distances than Baseline because Baseline does not attempt to reduce the cross-camera discrepancy. This demonstrates that the proposed MCAL can make each camera aligned as much as possible. 
Besides, this experiment also shows that the OCE-based scheme achieves the smallest distance, showing its best capability in reducing the discrepancy across cameras on Market1501, DukeMTMC-reID and MSMT17. This is attributed to the proposed special camera-label assigning scheme. Additionally, we visualize the data distributions obtained by the feature representation from Baseline, ACAN-GRL and ACAN-OCE in Fig.~\ref{fig7}. The result further illustrates the efficacy of the camera-alignment module in the proposed framework. As seen, compared with the baseline method, our method makes samples closer in the feature space (Fig.~\ref{fig7} (a) vs. Fig.~\ref{fig7} (e)  and Fig.~\ref{fig7} (b) vs. Fig.~\ref{fig7} (f)), \ie, there are more unoccupied regions in Fig.~\ref{fig7} (e) and Fig.~\ref{fig7} (f) than Fig.~\ref{fig7} (a) and Fig.~\ref{fig7}(b), respectively. This smaller feature space illustrates that the data-distribution discrepancy across different cameras is smaller (\ie, $d_{\mathrm{inter-camera}}$ is a smaller value). Besides, on MSMT17, the blue dots have a high concentration in a small region in Fig.~\ref{fig7} (b) (\ie, baseline), which means samples from this camera could form a specific sub-region of the feature space. Differently, these blue dots in Fig.~\ref{fig7} (f) (\ie, our method) are well scattered in the feature space, which shows our method can achieve better alignment of the samples from different cameras than the baseline.
Based on the above results and analysis, we can also observe that \textit{the OCE scheme can well achieve camera alignment over the GRL scheme, while this does not mean that using OCE can obtain better performance than the GRL scheme in our unsupervised cross-camera person Re-ID.} For example, on Market1501, although using the OCE scheme can better reduce the data distribution discrepancy than the GRL scheme in Table~\ref{tab07}, the latter can obtain better Re-ID performance than the former in Table~\ref{tab05}. The main reason has been discussed in Section~\ref{SEC:AC-2}.

\begin{figure}
\centering
\subfigure[Market (ACAN-GRL)]{
\includegraphics[width=4cm]{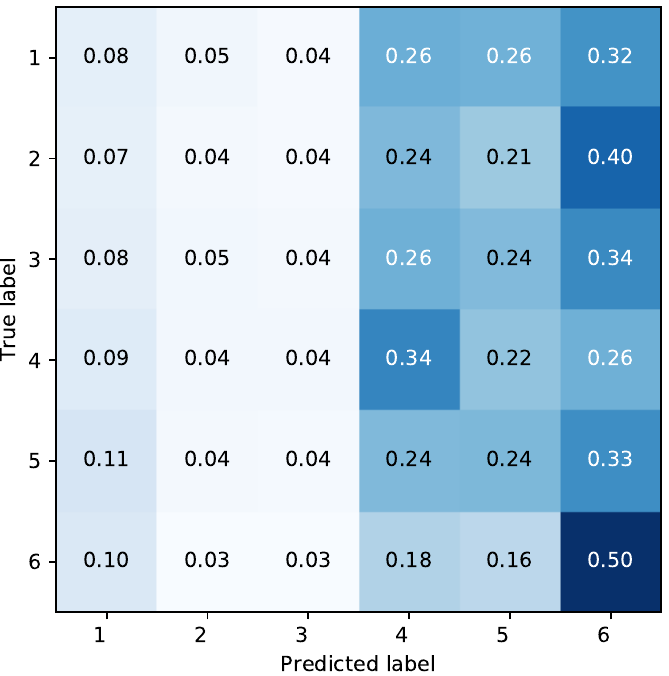}
}
\subfigure[Market (ACAN-OCE)]{
\includegraphics[width=4cm]{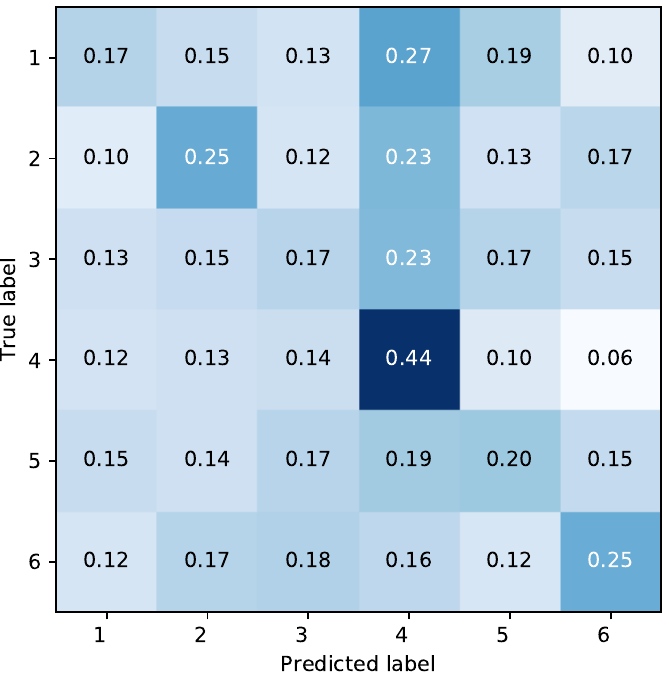}
}
\subfigure[Duke (ACAN-GRL)]{
\includegraphics[width=4cm]{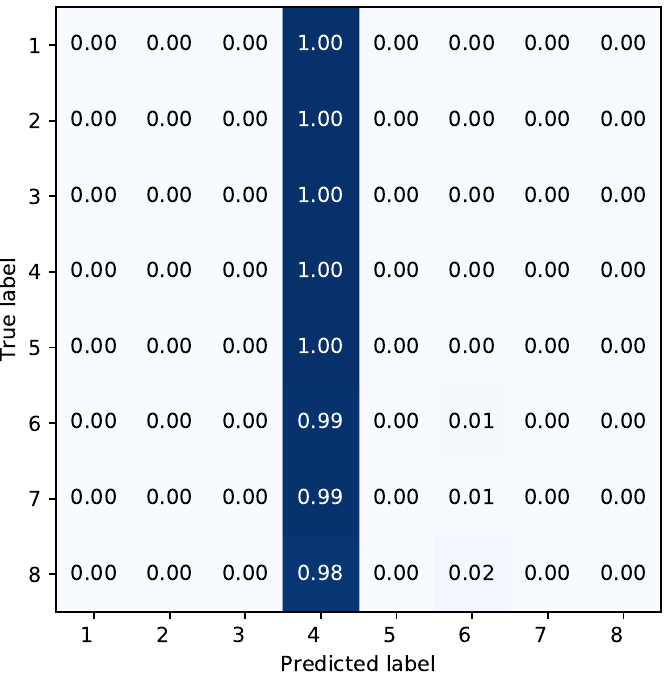}
}
\subfigure[Duke (ACAN-OCE)]{
\includegraphics[width=4cm]{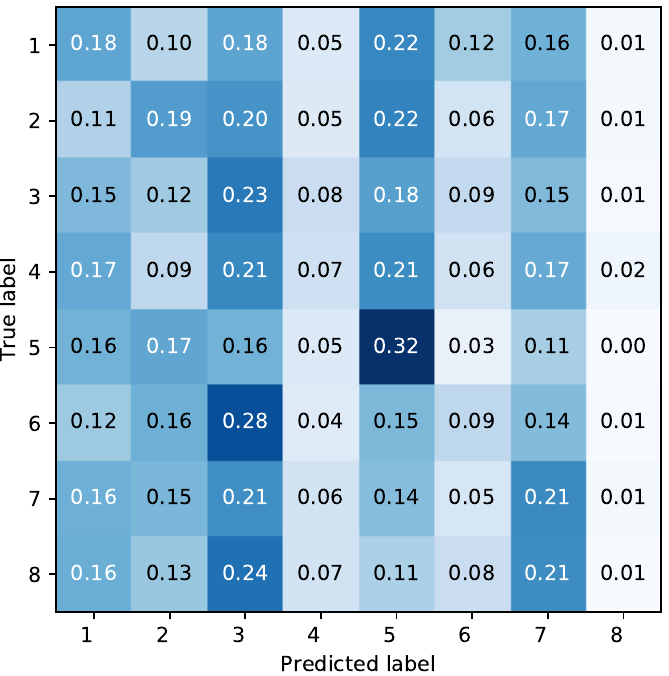}
}
\subfigure[MSMT (ACAN-GRL)]{
\includegraphics[width=4cm]{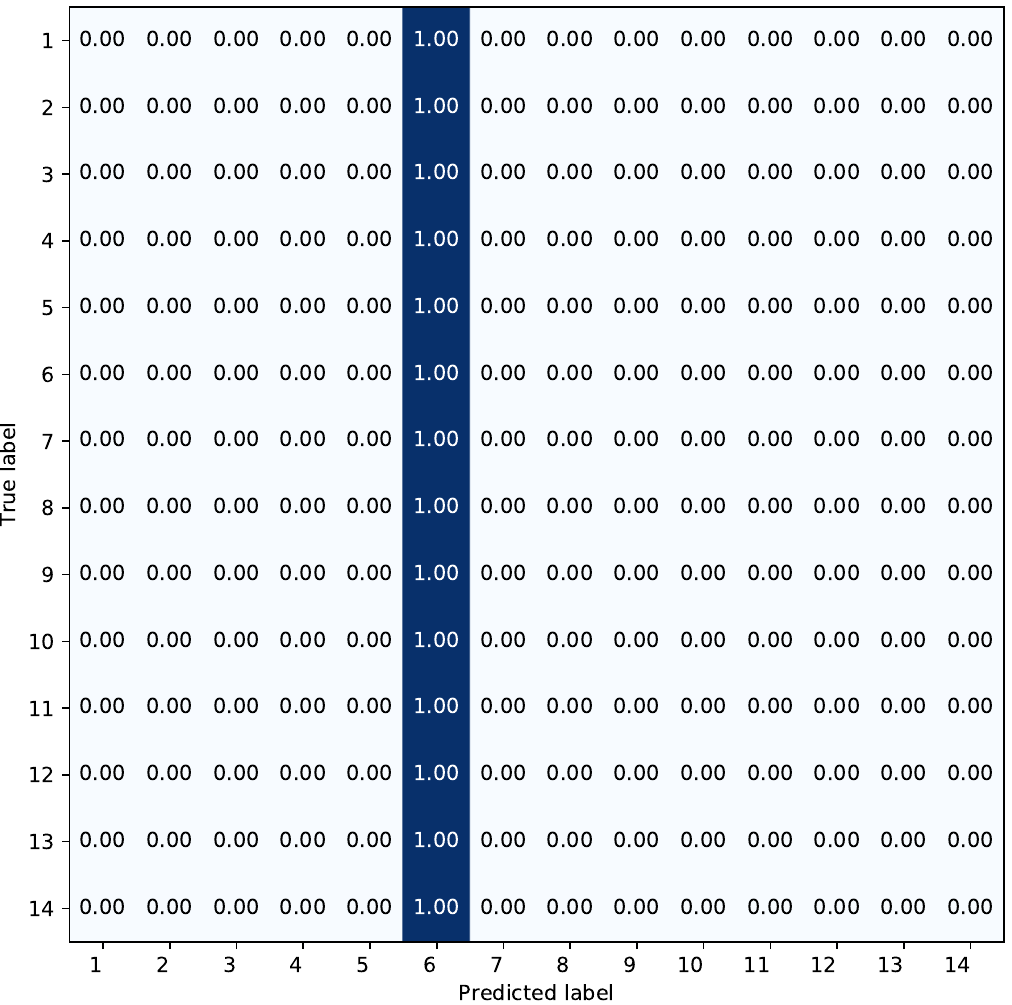}
}
\subfigure[MSMT (ACAN-OCE)]{
\includegraphics[width=4cm]{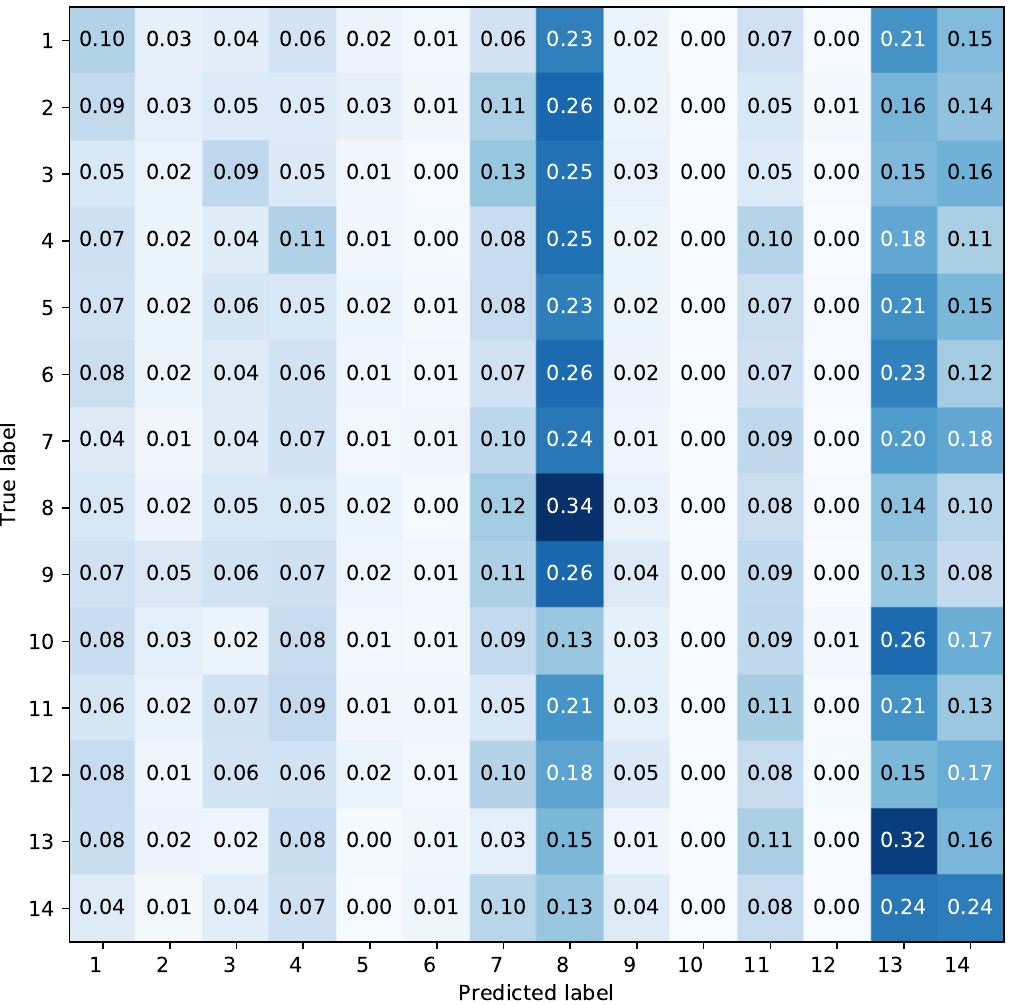}
}
\caption{Visualization of the camera confusion matrices on Market1501 (Market), DukeMTMC-reID (Duke) and MSMT17 (MSMT). In detail, (a) and (b) are six different cameras on Market1501; (c) and (d) are eight different cameras on DukeMTMC-reID. (e) and (f) are fourteen different cameras on MSMT17. Best viewed by horizontal contrast.}
\label{fig8}
\vspace*{-15pt}
\end{figure}

\textbf{Camera Confusion Matrix.}
To further compare the difference between the GRL-based and OCE-based schemes, we visualize the camera confusion matrices of GRL and OCE on Market1501 (Market), DukeMTMC-reID (Duke) and MSMT17 (MSMT), as shown in Fig.~\ref{fig8}. As seen, the OCE-based scheme approximately confuses across multiple cameras, while the GRL-based scheme could mix local cameras, especially for these datasets with more cameras, such as DukeMTMC-reID collected from 8 cameras and MSMT17 captured from 15 cameras. This main reason is that maximizing the discriminator by GRL is approximate to freely assigning the camera label of an image except for the camera which this image belongs to. In addition, the confusion matrices of the OCE-based scheme are consistent with the analysis in \textit{Corollary $1$} of Section~\ref{SEC:AC-2}, i.e., if the discriminator outputs equal probability for all cameras, all images can be mapped into a shared space. Compared with GRL, the confusion matrices of OCE in Fig.~\ref{fig8} tend to achieve this goal. As seen in Fig.~\ref{fig8}, on Duke, the OCE scheme is more balanced to output the predicted probability for each camera class, while the GRL scheme almost assigns all images into one camera. This further confirms the efficacy of the OCE scheme. 
\begin{table}[htbp]
  \centering
  \caption{Evaluation of the ACE scheme and the OCE scheme on three image datasets (i.e., Market1501, DukeMTMC-reID and MSMT17) and two video datasets (i.e., DukeMTMC-SI-Tracklet (Duke-T) and MARS). The best performance is shown in \textbf{bold}.}
    \begin{tabular}{|c|c|cccc|}
    \toprule
    \midrule
     Dataset & Method & mAP   & Rank-1 & Rank-5 & Rank-10 \\
    \midrule
    \multirow{2}[1]{*}{Market1501} & ACE  & 39.8  & 64.0  & 80.6  & 86.2  \\
\cmidrule{2-6}          & OCE (ours) & \textbf{47.7} & \textbf{72.2} & \textbf{86.3} & \textbf{90.4} \\
    \midrule
    \midrule
    \multirow{2}[1]{*}{Duke} & ACE  & 16.0  & 37.3  & 52.0  & 59.4   \\
\cmidrule{2-6}          & OCE (ours) & \textbf{45.1} & \textbf{67.6} & \textbf{81.2} & \textbf{85.2}  \\
    \midrule
    \midrule
    \multirow{2}[1]{*}{MSMT17} & ACE  & 3.9   & 13.0  & 23.3  & 29.1  \\
\cmidrule{2-6}          & OCE (ours) & \textbf{12.6} & \textbf{33.0} & \textbf{48.0} & \textbf{54.7}\\
    \midrule
    \midrule
    \multirow{2}[1]{*}{MARS} & ACE  & 40.9  & 53.9  & 71.4  & 77.1   \\
\cmidrule{2-6}          & OCE (ours) & \textbf{47.5}  & \textbf{57.7}  & \textbf{75.1}  & \textbf{79.9} \\
    \midrule
    \midrule
    \multirow{2}[1]{*}{Duke-T} & ACE  & 26.5  & 38.7  & 53.8  & 60.0  \\
\cmidrule{2-6}          & OCE (ours) & \textbf{40.3} & \textbf{50.4} & \textbf{68.0} & \textbf{74.1}  \\
    \bottomrule
    \end{tabular}%
  \label{tab06}%
\end{table}%



\begin{table}[htbp]
  \centering
  \caption{The discrepancy of data distribution across different cameras on Market1501, DukeMTMC-reID and MSMT17. Note that a smaller value indicates better performance in this table. The best performance is shown in \textbf{bold}}
    \begin{tabular}{|c|c|c|c|}
    \toprule
     \midrule
    Method & Market1501 & ~~Duke~~  & MSMT17 \\
    \midrule
    Baseline ($\times$100) & 9.69  & 8.33  & 10.85 \\
    GRL  ($\times$100) & 3.54  & 6.21  & 8.59 \\
    OCE (ours)  ($\times$100) & \textbf{2.36}  & \textbf{2.45}  & \textbf{2.16} \\
    \bottomrule
    \end{tabular}%
  \label{tab07}%
\end{table}%

\begin{figure}[h]
\centering
\subfigure[Market1501]{
\includegraphics[width=4.1cm]{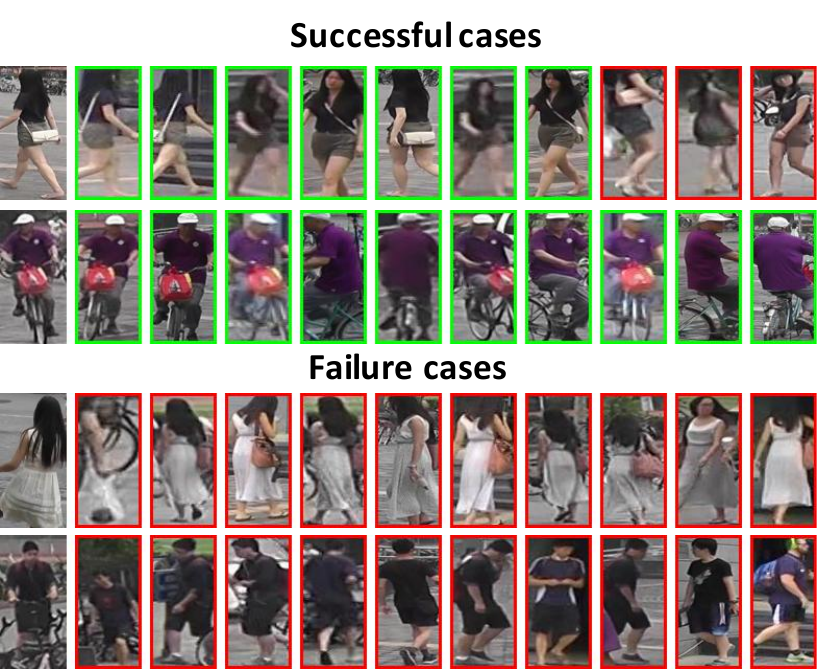}
}
\subfigure[DukeMTMC-reID]{
\includegraphics[width=4.1cm]{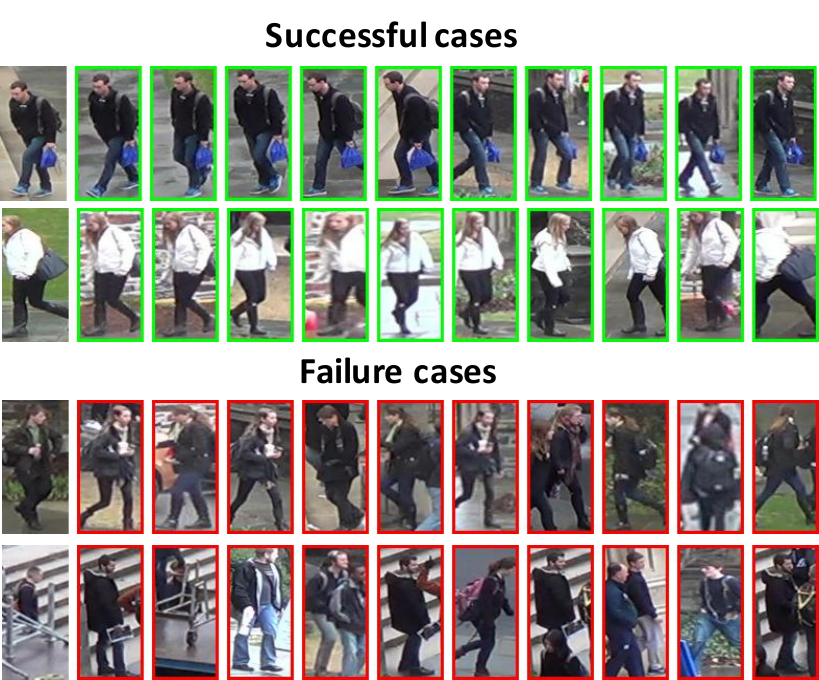}
}
\caption{The visual results of successful cases and failure cases.}
\label{fig_r2}
\vspace{-20pt}
\end{figure}

\textbf{Visual Examples.}
We add some visual examples from successful and failure cases, as shown in Fig.~\ref{fig_r2}. As seen, our multi-camera adversarial learning can effectively mitigate the issue of the pose difference. However, since our method cannot introduce some discriminative information (\eg, pseudo labels), it could fail when these persons in query images and gallery images have the same color cloth yet different identities. For this problem, we will further propose a new solution in future work.

\begin{figure}[h]
\centering
\subfigure[\#camera]{
\includegraphics[width=4cm]{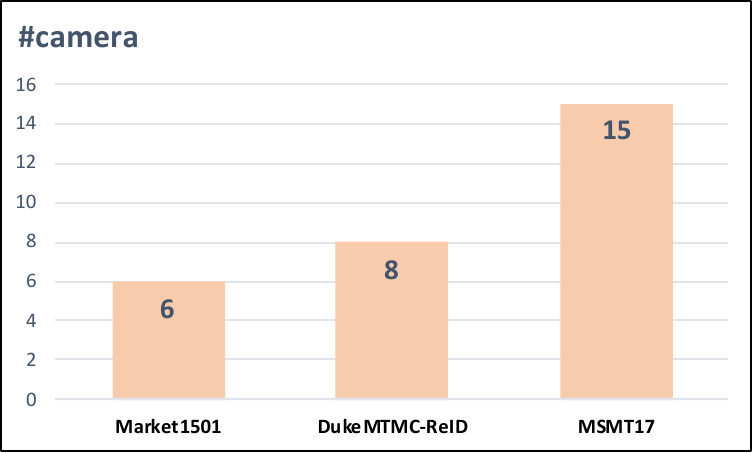}
}
\subfigure[\#image]{
\includegraphics[width=4cm]{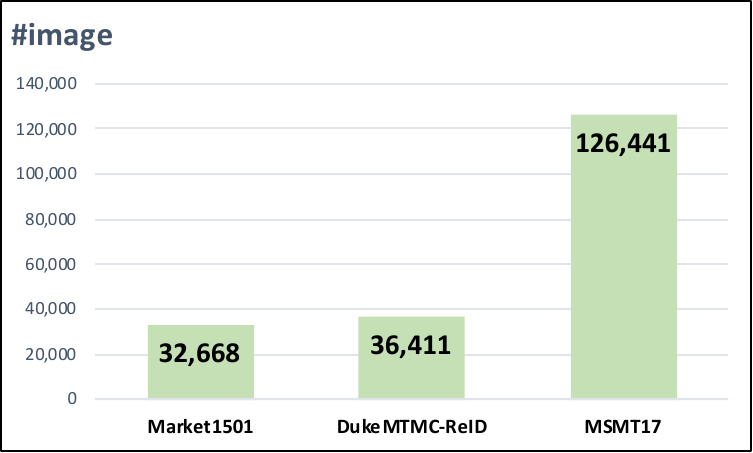}
}
\caption{The numbers of cameras and images on three benchmark datasets.}
\label{fig_r3}
\vspace*{-15pt}
\end{figure}

\textbf{Comparison with Fully Unsupervised Methods.}
Both our method and the fully unsupervised methods are the target-only Re-ID methods~\cite{ding2020adaptive}, while there are two differences between them as follows: 1) compared with the fully unsupervised methods, our method uses the intra-camera labels gained readily by tracking algorithms and few annotations, thus it can help our method to achieve better performance on the large-scale datasets; 2) Our method and fully unsupervised methods adopt different techniques to handle the unlabeled case. Specifically, these existing fully unsupervised SOTA methods mine discriminative information to train the model, while our method resorts to mitigate the discrepancy of data distribution across cameras, which is a simple yet effective scheme.
To further confirm the advantage of our setting and method, we take two fully unsupervised SOTA methods (\ie, BUC~\cite{lin2019bottom} and AE~\cite{ding2020adaptive}) to compare with our method, as reported in Table~\ref{tab-v2-01}. Specifically, BUC~\cite{lin2019bottom} is a bottom-up clustering approach to jointly optimize a convolutional neural network (CNN) and the relationship among the individual samples. 
AE~\cite{ding2020adaptive} is an adaptive exploration method to address the domain-shift problem for Re-ID in an unsupervised manner, which can maximize distances between all person images and minimize distances between similar person images.
 Compare to these methods, our method can obtain significantly better performance on large-scale datasets, such as Duke and MSMT17, which is attributed to the proposed camera adversarial learning scheme and our setting. As shown in Fig.~\ref{fig_r3},  Duke and MSMT17 include more cameras and images than Market1501, which can better reveal the superiority of our method.  In addition, because more cameras bring an increased number of different data distributions, a dataset with fewer cameras is usually easier to handle than a dataset with more cameras. For the pseudo-label-based method (\eg, AE~\cite{ding2020adaptive}), when there are only a small number of cameras for a dataset (\eg, Market1501), producing accurate pseudo-labels will become easier. In this case, the pseudo-label-based method (\eg, AE~\cite{ding2020adaptive}) can achieve better performance than our method.

\begin{table}[htbp]
  \centering
  \caption{Comparison to fully unsupervised SOTA re-ID methods.}
 \begin{tabular}{|C{2.2cm}|C{0.3cm}C{0.9cm}|C{0.3cm}C{0.9cm}|C{0.3cm}C{0.9cm}|}
     \toprule
    \midrule
    \multirow{2}[1]{*}{Method}  & \multicolumn{2}{c|}{Duke} & \multicolumn{2}{c|}{MSMT17} & \multicolumn{2}{c|}{Market1501}\\
\cmidrule{2-7}          & mAP   & Rank-1 & mAP   & Rank-1 & mAP   & Rank-1 \\
    \midrule
    BUC~\cite{lin2019bottom}    & 27.5  & 47.4  & 3.4   & 11.5 & 38.3  & 66.2 \\
    AE~\cite{ding2020adaptive}    & 39.0  & 63.2  & 8.5   & 26.6 & \textbf{54.0} & \textbf{77.5}\\
    \midrule
    ACAN-GRL  (ours)  & \textbf{46.6} & 65.1  & 11.2  & 27.1 & 50.6  & 73.3\\
    ACAN-OCE  (ours)  & 45.1  & \textbf{67.6} & \textbf{12.6} & \textbf{33.0} & 47.7  & 72.2 \\
    \bottomrule
    \end{tabular}%
  \label{tab-v2-01}%
  \vspace*{-10pt}
\end{table}%

\textbf{The Robustness of Our Method.}
To further demonstrate the robustness of our method, we conduct a new experiment by only retaining the images under one camera for 25\% or 50\% identities, \ie, 25\% or 50\% persons will only appear under one (but could be different) camera. \\
Meanwhile, the above step will only retain a part of the original images in the training set and this will certainly lead to an unfair comparison with the baseline, which is obtained with all of the 12936 training images on Market1501. As shown in Table~\ref{tab-v2-02},  ``single-25/50'' denotes the case in which 25/50\% persons will only appear under a single camera. We can see that the corresponding number of training samples decreases from the original 12936 to 10436 and 7895, respectively.\\
To avoid such an unfair comparison, we have to remove the effect caused by the different number of training images. Therefore, we recreate the original training set to ensure it has almost the same number of training images in each case as follows: \\
-~For ``single-25/50'' in Table~\ref{tab-v2-02}, we randomly select 25\% or 50\% persons from the 751 persons. For each selected person, we randomly choose one camera to retain all images of this person from it and remove all images of this person from the rest cameras.\\ 
-~For each corresponding ``multiple-all''\footnote{``multiple-all'' means that all persons appear under multiple cameras (i.e., \#camera $\geq$ 2).} in Table~\ref{tab-v2-02}, we randomly select one person from the 751 persons. If the chosen person appears under three or more cameras (i.e., \#camera $\geq$ 3), we will randomly select one camera to remove all images of this person from it. We iteratively conduct the above two steps until the number of retained images is almost equal to the number of training images obtained in the above ``single-25/50'' cases. In doing so, we can largely remove the effect caused by the different number of training images when conducting comparison. The statistics of the created training set in each case are reported in Tables~\ref{tab-v2-02} and~\ref{tab-v2-03}. As seen, 1) the ``single-25/50'' setting and the corresponding ``multiple-all'' have almost the same number of training images in Tables~\ref{tab-v2-02}; 2) In Table~\ref{tab-v2-03}, the ``single-25/50'' setting includes multiple persons that only appear under one camera, while in the corresponding ``multiple-all'' setting, all persons are captured by multiple cameras (\ie, \#camera $\geq$ 2).\\
To accumulate statistics, we repeat the above process to create five training sets and report the results averaged on them in Table~\ref{tab-v2-02}. As seen, the proposed method works better in the setting of ``single-25/50'' than ``multiple-all'' in most cases\footnote{At the same time, it can be well expected that the proposed method in these settings cannot compete with the original setting in which the number of training images is much larger.}. This result demonstrates the robustness of the proposed method when a considerable number of persons only have images under a single camera.

\begin{table*}
  \centering
  \caption{Comparison between baseline and the proposed method in the cases of keeping images of only one camera for 25\% and 50\% identities. Note ``single-25'' denotes 25\% persons appear under one camera only. ``multiple-all'' means that all persons appear under multiple cameras (i.e., \#camera $\geq$ 2). ``original'' indicates the baseline using the original training setting on Market1501.}
    \begin{tabular}{|c|c|c|cc|cc|}
    \toprule
    \midrule
    \multirow{2}[1]{*}{Setting} & \multirow{2}[1]{*}{\#training image} & \multirow{2}[1]{*}{\#training ID} & \multicolumn{2}{c|}{ACAN-GRL} & \multicolumn{2}{c|}{ACAN-OCE} \\
\cmidrule{4-7}          &   &    & mAP   & Rank-1 & mAP   & Rank-1 \\
    \midrule
    singel-25 & 10436 & 751 & 37.3  & 57.9  & 38.4  & 62.6 \\
    mulitple-all & 10395 & 751 & 34.5  & 56.1  & 38.1  & 62.0 \\
    \midrule
    singel-50 & 7895& 751 & 32.5  & 53.3  & 35.5  & 59.9 \\
    mulitple-all & 7894 & 751 & 32.2  & 53.2  & 35.9  & 59.4 \\
    \bottomrule
    \midrule
   original & 12936 & 751 & 50.6 & 73.3 & 47.7 & 72.2 \\
    \bottomrule
    \end{tabular}%
  \label{tab-v2-02}%

\end{table*}%

\begin{table*}[htbp]
  \centering
  \caption{The statistics of the constructed training set in each case. \#ID-2CAM denotes the number of persons who appear under 2 cameras. Note that the averaged value over the five constituted training sets are given in this table.}
    \begin{tabular}{|c|cccccc|}
    \toprule
    \midrule
    Setting & \#ID-1CAM  & \#ID-2CAM    & \#ID-3CAM     & \#ID-4CAM      & \#ID-5CAM     & \#ID-6CAM \\
    \midrule
    single-25 & 188.0 & 36.4  & 81.2  & 156.2 & 226.6 & 62.6 \\
    multiple-all & 0.0   & 189.0 & 196.6 & 212.4 & 126.6 & 26.4 \\
    \midrule
    single-50 & 376.0 & 25.0  & 50.8  & 106.6 & 153.0 & 39.6 \\
    multiple-all & 0.0   & 396.0 & 238.6 & 95.0  & 19.4  & 2.0 \\
    \bottomrule
    \end{tabular}%
   \label{tab-v2-03}%
   \vspace*{-15pt}
\end{table*}%

\textbf{More Discussion.}
Since GRL is developed to align two different domains, for two more domains, if using GRL could not effectively reduce the domain shift, which has been analyzed in Section III-A of our paper. From this perspective, we put forward the OCE scheme to address this issue, thus the proposed OCE can achieve better alignment for multiple domains. However, as seen in the experimental results, despite the OCE scheme can better mitigate the domain shift when compared with GRL as shown in Table VII of our paper, the performance of OCE cannot consistently outperform GRL on all datasets. It is worth noting that our method can work well on a large dataset MSMT17, which includes more cameras and images than Duke and MSMT17 as shown in Fig.~\ref{fig_r3}\footnote{Video datasets (including MARS and DukeMTMC-SI-Tracklet) and image datasets (including Market1501 and Duke) are from the same scenario, thus they have the same number of cameras (\ie, 6 cameras and 8 cameras)}. Therefore, in our research, we got an observation that the OCE scheme can well achieve camera alignment over the GRL scheme, while this does not mean that using OCE can obtain better performance than the GRL scheme in Re-ID on the ``small-scale'' dataset with a few cameras. Therefore, in Re-ID research, we cannot only pursue to reduce the data-distribution discrepancy, but also consider the relationship of original data space for the small-scale dataset, thus it brings us a reminder in the further research.

\section{Conclusion}\label{s-conclusion}
In this paper, we focus on a new person re-identification task named unsupervised cross-camera person Re-ID. It only assumes the availability of label information within the images from the same camera but no inter-camera label information is provided. Based on the perspective of reducing the cross-camera data distribution, we propose a novel adversarial camera alignment network (ACAN) for the proposed person Re-ID task, which can map all images from different cameras into a shared subspace. To realize the camera alignment task, we put forward a multi-camera adversarial learning method which is achieved by two strategies including the GRL-based and OCE-based schemes and give the corresponding theoretical analysis. 
Extensive experiments show the superiority of our proposed ACAN and confirm the efficacy of all components in ACAN. Particularly, to deal with the proposed task, a more intuitive strategy is generating pseudo-labels to explore the relationship between cross-camera samples. In future work, we will integrate the pseudo-label-based methods to further enhance the generalization ability of person Re-ID with the within-camera labeled information in real-world applications.


%
%

\ifCLASSOPTIONcaptionsoff
  \newpage
\fi

\bibliographystyle{IEEEtran}
\bibliography{sigproc}

\begin{thebibliography}{10}
\providecommand{\url}[1]{#1}
\csname url@samestyle\endcsname
\providecommand{\newblock}{\relax}
\providecommand{\bibinfo}[2]{#2}
\providecommand{\BIBentrySTDinterwordspacing}{\spaceskip=0pt\relax}
\providecommand{\BIBentryALTinterwordstretchfactor}{4}
\providecommand{\BIBentryALTinterwordspacing}{\spaceskip=\fontdimen2\font plus
\BIBentryALTinterwordstretchfactor\fontdimen3\font minus
  \fontdimen4\font\relax}
\providecommand{\BIBforeignlanguage}[2]{{%
\expandafter\ifx\csname l@#1\endcsname\relax
\typeout{** WARNING: IEEEtran.bst: No hyphenation pattern has been}%
\typeout{** loaded for the language `#1'. Using the pattern for}%
\typeout{** the default language instead.}%
\else
\language=\csname l@#1\endcsname
\fi
#2}}
\providecommand{\BIBdecl}{\relax}
\BIBdecl

\bibitem{DBLP:journals/tmm/ChenLLCH11}
K.~Chen, C.~Lai, P.~Lee, C.~Chen, and Y.~Hung, ``Adaptive learning for target
  tracking and true linking discovering across multiple non-overlapping
  cameras,'' \emph{IEEE Trans. Multimedia (TMM)}, vol.~13, no.~4, pp. 625--638,
  2011.

\bibitem{DBLP:journals/tip/ZhangLZZZ15}
R.~Zhang, L.~Lin, R.~Zhang, W.~Zuo, and L.~Zhang, ``Bit-scalable deep hashing
  with regularized similarity learning for image retrieval and person
  re-identification,'' \emph{IEEE Trans. Image Processing (TIP)}, vol.~24,
  no.~12, pp. 4766--4779, 2015.

\bibitem{DBLP:journals/tmm/Wang0TL13}
X.~Wang, T.~Zhang, D.~Tretter, and Q.~Lin, ``Personal clothing retrieval on
  photo collections by color and attributes,'' \emph{IEEE Trans. Multimedia
  (TMM)}, vol.~15, no.~8, pp. 2035--2045, 2013.

\bibitem{DBLP:journals/tifs/MaJZTP20}
F.~Ma, X.~Jing, X.~Zhu, Z.~Tang, and Z.~Peng, ``True-color and grayscale video
  person re-identification,'' \emph{IEEE Trans. Information Forensics and
  Security (TIFS)}, vol.~15, pp. 115--129, 2020.

\bibitem{DBLP:journals/tifs/YeL0Y20}
M.~Ye, X.~Lan, Z.~Wang, and P.~C. Yuen, ``Bi-directional center-constrained
  top-ranking for visible thermal person re-identification,'' \emph{IEEE Trans.
  Information Forensics and Security (TIFS)}, vol.~15, pp. 407--419, 2020.

\bibitem{DBLP:conf/cvpr/XiaoLOW16}
T.~Xiao, H.~Li, W.~Ouyang, and X.~Wang, ``Learning deep feature representations
  with domain guided dropout for person re-identification,'' in \emph{IEEE
  Conference on Computer Vision and Pattern Recognition (CVPR)}, 2016, pp.
  1249--1258.

\bibitem{DBLP:conf/aaai/ChenCZH17}
W.~Chen, X.~Chen, J.~Zhang, and K.~Huang, ``A multi-task deep network for
  person re-identification,'' in \emph{Proceedings of the Thirty-First
  Conference on Artificial Intelligence (AAAI)}, 2017, pp. 3988--3994.

\bibitem{sun2018beyond}
Y.~Sun, L.~Zheng, Y.~Yang, Q.~Tian, and S.~Wang, ``Beyond part models: Person
  retrieval with refined part pooling (and a strong convolutional baseline),''
  in \emph{European Conference on Computer Vision (ECCV)}, 2018, pp. 501--518.

\bibitem{zheng2018re}
M.~Zheng, S.~Karanam, Z.~Wu, and R.~J. Radke, ``Re-identification with
  consistent attentive siamese networks,'' in \emph{IEEE Conference on Computer
  Vision and Pattern Recognition (CVPR)}, 2019, pp. 5735--5744.

\bibitem{zheng2019pose}
L.~Zheng, Y.~Huang, H.~Lu, and Y.~Yang, ``Pose invariant embedding for deep
  person re-identification,'' \emph{IEEE Trans. on Image Processing (TIP)},
  vol.~28, no.~9, pp. 4500--4509, 2019.

\bibitem{wu2019cross}
L.~Wu, R.~Hong, Y.~Wang, and M.~Wang, ``Cross-entropy adversarial view
  adaptation for person re-identification,'' \emph{IEEE Trans. on Circuits and
  Systems for Video Technology (TCSVT)}, vol.~30, no.~7, pp. 2081--2092, 2019.

\bibitem{fan2018unsupervised}
H.~Fan, L.~Zheng, C.~Yan, and Y.~Yang, ``Unsupervised person re-identification:
  clustering and fine-tuning,'' \emph{ACM Trans. on Multimedia Computing,
  Communications, and Applications (TOMM)}, vol.~14, no.~4, pp. 83:1--83:18,
  2018.

\bibitem{zhong2018generalizing}
Z.~Zhong, L.~Zheng, S.~Li, and Y.~Yang, ``Generalizing a person retrieval model
  hetero-and homogeneously,'' in \emph{European Conference on Computer Vision
  (ECCV)}, 2018, pp. 176--192.

\bibitem{Bak_2018_ECCV}
S.~Bak, P.~Carr, and J.-F. Lalonde, ``Domain adaptation through synthesis for
  unsupervised person re-identification,'' in \emph{European Conference on
  Computer Vision (ECCV)}, 2018, pp. 193--209.

\bibitem{wang2018transferable}
J.~Wang, X.~Zhu, S.~Gong, and W.~Li, ``Transferable joint attribute-identity
  deep learning for unsupervised person re-identification,'' in \emph{IEEE
  Conference on Computer Vision and Pattern Recognition (CVPR)}, 2018, pp.
  2275--2284.

\bibitem{lv2018unsupervised}
J.~Lv, W.~Chen, Q.~Li, and C.~Yang, ``Unsupervised cross-dataset person
  re-identification by transfer learning of spatial-temporal patterns,'' in
  \emph{IEEE Conference on Computer Vision and Pattern Recognition (CVPR)},
  2018, pp. 7948--7956.

\bibitem{dehghan2015gmmcp}
A.~Dehghan, S.~Modiri~Assari, and M.~Shah, ``Gmmcp tracker: Globally optimal
  generalized maximum multi clique problem for multiple object tracking,'' in
  \emph{IEEE Conference on Computer Vision and Pattern Recognition (CVPR)},
  2015, pp. 4091--4099.

\bibitem{maksai2017non}
A.~Maksai, X.~Wang, F.~Fleuret, and P.~Fua, ``Non-markovian globally consistent
  multi-object tracking,'' in \emph{International Conference on Computer Vision
  (ICCV)}, 2017, pp. 2563--2573.

\bibitem{DBLP:conf/bmvc/LinLLK18}
S.~Lin, H.~Li, C.~Li, and A.~C. Kot, ``Multi-task mid-level feature alignment
  network for unsupervised cross-dataset person re-identification,'' in
  \emph{British Machine Vision Conference (BMVC)}, 2018.

\bibitem{DBLP:journals/corr/abs-1904-01308}
G.~Delorme, X.~Alameda{-}Pineda, S.~Lathuili{\`{e}}re, and R.~Horaud, ``Camera
  adversarial transfer for unsupervised person re-identification,''
  \emph{arXiv}, 2019.

\bibitem{qi2019novel}
L.~Qi, L.~Wang, J.~Huo, L.~Zhou, Y.~Shi, and Y.~Gao, ``A novel unsupervised
  camera-aware domain adaptation framework for person re-identification,'' in
  \emph{International Conference on Computer Vision (ICCV)}, 2019, pp.
  8079--8088.

\bibitem{DBLP:conf/cvpr/HeZRS16}
K.~He, X.~Zhang, S.~Ren, and J.~Sun, ``Deep residual learning for image
  recognition,'' in \emph{IEEE Conference on Computer Vision and Pattern
  Recognition (CVPR)}, 2016, pp. 770--778.

\bibitem{DBLP:conf/cvpr/DengDSLL009}
J.~Deng, W.~Dong, R.~Socher, L.~Li, K.~Li, and F.~Li, ``Imagenet: {A}
  large-scale hierarchical image database,'' in \emph{IEEE Conference on
  Computer Vision and Pattern Recognition (CVPR)}, 2009, pp. 248--255.

\bibitem{maaten2008visualizing}
L.~v.~d. Maaten and G.~Hinton, ``Visualizing data using t-sne,'' \emph{Journal
  of Machine Learning Research (JMLR)}, vol.~9, no.~11, 2008.

\bibitem{liao2015person}
S.~Liao, Y.~Hu, X.~Zhu, and S.~Z. Li, ``Person re-identification by local
  maximal occurrence representation and metric learning,'' in \emph{IEEE
  conference on computer vision and pattern recognition (CVPR)}, 2015, pp.
  2197--2206.

\bibitem{DBLP:conf/iccv/ZhengSTWWT15}
L.~Zheng, L.~Shen, L.~Tian, S.~Wang, J.~Wang, and Q.~Tian, ``Scalable person
  re-identification: {A} benchmark,'' in \emph{International Conference on
  Computer Vision (ICCV)}, 2015, pp. 1116--1124.

\bibitem{DBLP:conf/cvpr/PengXWPGHT16}
P.~Peng, T.~Xiang, Y.~Wang, M.~Pontil, S.~Gong, T.~Huang, and Y.~Tian,
  ``Unsupervised cross-dataset transfer learning for person
  re-identification,'' in \emph{IEEE Conference on Computer Vision and Pattern
  Recognition (CVPR)}, 2016, pp. 1306--1315.

\bibitem{DBLP:journals/tcsv/WangZLZ16}
X.~Wang, W.~Zheng, X.~Li, and J.~Zhang, ``Cross-scenario transfer person
  re-identification,'' \emph{IEEE Trans. Circuits Syst. Video Techn. (TCSVT)},
  vol.~26, no.~8, pp. 1447--1460, 2016.

\bibitem{DBLP:journals/tip/MaLYL15}
A.~J. Ma, J.~Li, P.~C. Yuen, and P.~Li, ``Cross-domain person re-identification
  using domain adaptation ranking svms,'' \emph{IEEE Trans. Image Processing
  (TIP)}, vol.~24, no.~5, pp. 1599--1613, 2015.

\bibitem{qi2018unsupervised}
L.~Qi, J.~Huo, X.~Fan, Y.~Shi, and Y.~Gao, ``Unsupervised joint subspace and
  dictionary learning for enhanced cross-domain person re-identification,''
  \emph{IEEE Journal of Selected Topics in Signal Processing (JSTSP)}, vol.~12,
  no.~6, pp. 1263--1275, 2018.

\bibitem{DBLP:conf/iccv/YuWZ17}
H.~Yu, A.~Wu, and W.~Zheng, ``Cross-view asymmetric metric learning for
  unsupervised person re-identification,'' in \emph{International Conference on
  Computer Vision (ICCV)}, 2017, pp. 994--1002.

\bibitem{wei2018person}
L.~Wei, S.~Zhang, W.~Gao, and Q.~Tian, ``Person transfer gan to bridge domain
  gap for person re-identification,'' in \emph{IEEE Conference on Computer
  Vision and Pattern Recognition (CVPR)}, 2018, pp. 79--88.

\bibitem{deng2018image}
W.~Deng, L.~Zheng, G.~Kang, Y.~Yang, Q.~Ye, and J.~Jiao, ``Image-image domain
  adaptation with preserved self-similarity and domain-dissimilarity for person
  reidentification,'' in \emph{IEEE Conference on Computer Vision and Pattern
  Recognition (CVPR)}, 2018, pp. 994--1003.

\bibitem{zhong2019invariance}
Z.~Zhong, L.~Zheng, Z.~Luo, S.~Li, and Y.~Yang, ``Invariance matters: Exemplar
  memory for domain adaptive person re-identification,'' in \emph{IEEE
  Conference on Computer Vision and Pattern Recognition (CVPR)}, 2019, pp.
  598--607.

\bibitem{kodirov2016person}
E.~Kodirov, T.~Xiang, Z.~Fu, and S.~Gong, ``Person re-identification by
  unsupervised $l_{1}$ graph learning,'' in \emph{European Conference on
  Computer Vision (ECCV)}, 2016, pp. 178--195.

\bibitem{khan2016unsupervised}
F.~M. Khan and F.~Bremond, ``Unsupervised data association for metric learning
  in the context of multi-shot person re-identification,'' in
  \emph{International Conference on Advanced Video and Signal Based
  Surveillance (AVSS)}, 2016, pp. 256--262.

\bibitem{ye2018robust}
M.~Ye, X.~Lan, and P.~C. Yuen, ``Robust anchor embedding for unsupervised video
  person re-identification in the wild,'' in \emph{European Conference on
  Computer Vision (ECCV)}, 2018, pp. 176--193.

\bibitem{liu2017stepwise}
Z.~Liu, D.~Wang, and H.~Lu, ``Stepwise metric promotion for unsupervised video
  person re-identification,'' in \emph{International Conference on Computer
  Vision (ICCV)}, 2017, pp. 2448--2457.

\bibitem{ye2017dynamic}
M.~Ye, A.~J. Ma, L.~Zheng, J.~Li, and P.~C. Yuen, ``Dynamic label graph
  matching for unsupervised video re-identification,'' in \emph{International
  Conference on Computer Vision (ICCV)}, 2017, pp. 5152--5160.

\bibitem{DBLP:conf/bmvc/ChenZG18}
Y.~Chen, X.~Zhu, and S.~Gong, ``Deep association learning for unsupervised
  video person re-identification,'' in \emph{British Machine Vision Conference
  (BMVC)}, 2018, p.~48.

\bibitem{Li_2018_ECCV}
M.~Li, X.~Zhu, and S.~Gong, ``Unsupervised person re-identification by deep
  learning tracklet association,'' in \emph{European Conference on Computer
  Vision (ECCV)}, 2018, pp. 772--788.

\bibitem{li2019unsupervised}
M.~Li, X.~Zhu, S.~Gong, and S.~Gong, ``Unsupervised tracklet person
  re-identification,'' \emph{IEEE Trans. on Pattern Analysis and Machine
  Intelligence (TPAMI)}, vol.~42, no.~7, pp. 1770--1782, 2020.

\bibitem{DBLP:conf/icml/LongC0J15}
M.~Long, Y.~Cao, J.~Wang, and M.~I. Jordan, ``Learning transferable features
  with deep adaptation networks,'' in \emph{International Conference on Machine
  Learning (ICML)}, 2015, pp. 97--105.

\bibitem{NIPS2016_6110}
M.~Long, H.~Zhu, J.~Wang, and M.~I. Jordan, ``Unsupervised domain adaptation
  with residual transfer networks,'' in \emph{Advances in Neural Information
  Processing Systems (NIPS)}, 2016, pp. 136--144.

\bibitem{DBLP:journals/corr/ZhangYCW15}
X.~Zhang, F.~X. Yu, S.~Chang, and S.~Wang, ``Deep transfer network:
  Unsupervised domain adaptation,'' \emph{arXiv}, 2015.

\bibitem{DBLP:journals/corr/TzengHZSD14}
E.~Tzeng, J.~Hoffman, N.~Zhang, K.~Saenko, and T.~Darrell, ``Deep domain
  confusion: Maximizing for domain invariance,'' \emph{arXiv}, 2014.

\bibitem{DBLP:conf/eccv/GhifaryKZBL16}
M.~Ghifary, W.~B. Kleijn, M.~Zhang, D.~Balduzzi, and W.~Li, ``Deep
  reconstruction-classification networks for unsupervised domain adaptation,''
  in \emph{European Conference on Computer Vision (ECCV)}, 2016, pp. 597--613.

\bibitem{DBLP:conf/nips/BousmalisTSKE16}
K.~Bousmalis, G.~Trigeorgis, N.~Silberman, D.~Krishnan, and D.~Erhan, ``Domain
  separation networks,'' in \emph{Advances in Neural Information Processing
  Systems (NIPS)}, 2016, pp. 343--351.

\bibitem{DBLP:conf/icml/GaninL15}
Y.~Ganin and V.~S. Lempitsky, ``Unsupervised domain adaptation by
  backpropagation,'' in \emph{International Conference on Machine Learning
  (ICML)}, 2015, pp. 1180--1189.

\bibitem{DBLP:journals/corr/abs-1803-09210}
J.~Zhang, Z.~Ding, W.~Li, and P.~Ogunbona, in \emph{IEEE Conference on Computer
  Vision and Pattern Recognition (CVPR)}, 2018, pp. 8156--8164.

\bibitem{DBLP:conf/cvpr/TzengHSD17}
E.~Tzeng, J.~Hoffman, K.~Saenko, and T.~Darrell, ``Adversarial discriminative
  domain adaptation,'' in \emph{IEEE Conference on Computer Vision and Pattern
  Recognition (CVPR)}, 2017.

\bibitem{zhao2018adversarial}
H.~Zhao, S.~Zhang, G.~Wu, J.~M. Moura, J.~P. Costeira, and G.~J. Gordon,
  ``Adversarial multiple source domain adaptation,'' in \emph{Advances in
  Neural Information Processing Systems (NIPS)}, 2018, pp. 8568--8579.

\bibitem{xu2018deep}
R.~Xu, Z.~Chen, W.~Zuo, J.~Yan, and L.~Lin, ``Deep cocktail network:
  Multi-source unsupervised domain adaptation with category shift,'' in
  \emph{IEEE Conference on Computer Vision and Pattern Recognition (CVPR)},
  2018, pp. 3964--3973.

\bibitem{schoenauer-sebag2018multidomain}
A.~Schoenauer-Sebag, L.~Heinrich, M.~Schoenauer, M.~Sebag, L.~Wu, and
  S.~Altschuler, ``Multi-domain adversarial learning,'' in \emph{International
  Conference on Learning Representations (ICLR)}, 2019.

\bibitem{hermans2017defense}
A.~Hermans, L.~Beyer, and B.~Leibe, ``In defense of the triplet loss for person
  re-identification,'' \emph{arXiv}, 2017.

\bibitem{goodfellow2014generative}
I.~Goodfellow, J.~Pouget-Abadie, M.~Mirza, B.~Xu, D.~Warde-Farley, S.~Ozair,
  A.~Courville, and Y.~Bengio, ``Generative adversarial nets,'' in
  \emph{Advances in neural information processing systems (NIPS)}, 2014, pp.
  2672--2680.

\bibitem{boyd2004convex}
S.~Boyd and L.~Vandenberghe, \emph{Convex optimization}.\hskip 1em plus 0.5em
  minus 0.4em\relax Cambridge university press, 2004.

\bibitem{wu2019few}
L.~Wu, Y.~Wang, H.~Yin, M.~Wang, and L.~Shao, ``Few-shot deep adversarial
  learning for video-based person re-identification,'' \emph{IEEE Trans. on
  Image Processing (TIP)}, vol.~29, pp. 1233--1245, 2019.

\bibitem{ghosh2018multi}
A.~Ghosh, V.~Kulharia, V.~P. Namboodiri, P.~H. Torr, and P.~K. Dokania,
  ``Multi-agent diverse generative adversarial networks,'' in \emph{IEEE
  Conference on Computer Vision and Pattern Recognition (CVPR)}, 2018, pp.
  8513--8521.

\bibitem{DBLP:conf/iccv/ZhuPIE17}
J.~Zhu, T.~Park, P.~Isola, and A.~A. Efros, ``Unpaired image-to-image
  translation using cycle-consistent adversarial networks,'' in
  \emph{International Conference on Computer Vision (ICCV)}, 2017, pp.
  2242--2251.

\bibitem{DBLP:conf/iccv/ZhengZY17}
Z.~Zheng, L.~Zheng, and Y.~Yang, ``Unlabeled samples generated by {GAN} improve
  the person re-identification baseline in vitro,'' in \emph{International
  Conference on Computer Vision (ICCV)}, 2017, pp. 3774--3782.

\bibitem{DBLP:conf/eccv/ZhengBSWSWT16}
L.~Zheng, Z.~Bie, Y.~Sun, J.~Wang, C.~Su, S.~Wang, and Q.~Tian, ``{MARS:} {A}
  video benchmark for large-scale person re-identification,'' in \emph{European
  Conference on Computer Vision (ECCV)}, 2016, pp. 868--884.

\bibitem{felzenszwalb2009object}
P.~F. Felzenszwalb, R.~B. Girshick, D.~McAllester, and D.~Ramanan, ``Object
  detection with discriminatively trained part-based models,'' \emph{IEEE
  Trans. on pattern analysis and machine intelligence (TPAMI)}, vol.~32, no.~9,
  pp. 1627--1645, 2010.

\bibitem{tian2019imitating}
J.~Tian, Z.~Teng, R.~Li, Y.~Li, B.~Zhang, and J.~Fan, ``Imitating targets from
  all sides: An unsupervised transfer learning method for person
  re-identification,'' \emph{arXiv}, 2019.

\bibitem{DBLP:journals/tifs/RenLGZL20}
C.~Ren, B.~Liang, P.~Ge, Y.~Zhai, and Z.~Lei, ``Domain adaptive person
  re-identification via camera style generation and label propagation,''
  \emph{IEEE Trans. Information Forensics and Security (TIFS)}, vol.~15, pp.
  1290--1302, 2020.

\bibitem{ding2020adaptive}
Y.~Ding, H.~Fan, M.~Xu, and Y.~Yang, ``Adaptive exploration for unsupervised
  person re-identification,'' \emph{ACM Trans. on Multimedia Computing,
  Communications, and Applications (TOMM)}, vol.~16, no.~1, pp. 1--19, 2020.

\bibitem{lin2019bottom}
Y.~Lin, X.~Dong, L.~Zheng, Y.~Yan, and Y.~Yang, ``A bottom-up clustering
  approach to unsupervised person re-identification,'' in \emph{AAAI Conference
  on Artificial Intelligence (AAAI)}, 2019, pp. 8738--8745.

\end{thebibliography}

\end{document}